\documentclass{article}

\usepackage{iclr2027_conference,times}
\usepackage[T1]{fontenc}

\usepackage{amsmath,amsfonts,bm}

\def\eqref#1{equation~\ref{#1}}

\def\1{\bm{1}}

\DeclareMathAlphabet{\mathsfit}{\encodingdefault}{\sfdefault}{m}{sl}
\SetMathAlphabet{\mathsfit}{bold}{\encodingdefault}{\sfdefault}{bx}{n}

\usepackage{graphicx}
\usepackage{booktabs}
\usepackage{multirow}
\usepackage[table]{xcolor}
\definecolor{nvsfirst}{HTML}{DFEFD5}
\definecolor{nvssecond}{HTML}{FFF4C2}
\definecolor{nvsthird}{HTML}{D9EAF2}
\usepackage{microtype}
\usepackage{url}
\usepackage{placeins}
\usepackage{float}
\usepackage{needspace}
\usepackage{hyperref}

\title{InfiniSplat v2: Analyzing and Improving Feed-Forward 3DGS from Any Views}

\iclrfinalcopy
\author{%
\begin{tabular}[t]{@{}l@{\hspace{0.7em}}l@{\hspace{0.7em}}l@{\hspace{0.7em}}l@{\hspace{0.7em}}l@{}}
\textbf{Yongzhen Hu}$^{1,2*}$ & \textbf{Hao Yu}$^{1*}$ & \textbf{Jiawei Wang}$^{1}$ & \textbf{Xianming Zeng}$^{2}$ & \textbf{Jianyun Xu}$^{2}$\\
\textbf{Xingliang Liu}$^{2}$ & \textbf{Ruizhen Hu}$^{3}$ & \textbf{Xiaowei Zhou}$^{1}$ & \textbf{Hujun Bao}$^{1}$ & \textbf{Sida Peng}$^{1,4\dagger}$
\end{tabular}\\[4pt]
\normalfont $^{1}$Zhejiang University\quad $^{2}$Hello TransTech\quad $^{3}$Shenzhen University\quad $^{4}$Causa Robotics\\
\normalfont $^{*}$Equal contribution\quad $^{\dagger}$Corresponding author
}

\begin{document}

\maketitle
\fancyhead{}
\renewcommand{\headrulewidth}{0pt}
\vspace{-30pt}

\begin{figure}[H]
    \centering
    \includegraphics[width=1.00\linewidth]{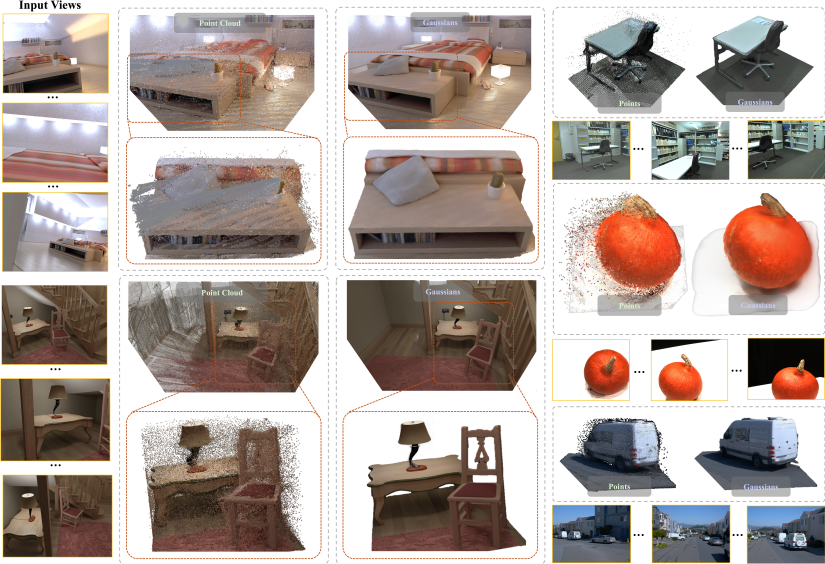}
\caption{From point clouds to Gaussian scenes.
InfiniSplat v2 retains point positions from a geometry foundation model
and learns Gaussian attributes for high-quality novel view rendering.
Left: input views, point clouds, and corresponding Gaussian renderings,
with enlarged regions outlined in orange. Right: additional examples comparing
point clouds with the resulting Gaussian scenes.}
    \label{fig:teaser}
\end{figure}

\vspace{-12pt}
\begin{abstract}
Feed-forward Gaussian reconstruction builds 3D scenes directly from input images
for novel view rendering, but learning Gaussian attributes from predicted geometry
for high-quality rendering under large viewpoint changes remains challenging.
We analyze limitations in Gaussian attribute learning and introduce InfiniSplat v2,
which improves rendering quality through architectural and training changes.
Our method decouples geometry and attribute prediction
with a frozen geometry branch, aligns predicted Gaussians using ground-truth
camera poses during training, and adopts large-angle target view supervision.
Across six datasets, the feed-forward outputs of the same trained model achieve
state-of-the-art mean LPIPS and DISTS among the evaluated methods in both
pose-conditioned and pose-free settings.
In the pose-conditioned setting, mean LPIPS and DISTS decrease by 25.8\% and
32.9\%, respectively, compared with the strongest evaluated baselines.
We further find that opacity learned with only image reconstruction
supervision suppresses geometrically inconsistent points.
Project page: \url{https://zju3dv.github.io/InfiniSplat2}.
\end{abstract}

\section{Introduction}
\label{sec:introduction}

Image-based 3D scene reconstruction aims to recover a renderable 3D scene
from input images to support novel view rendering for applications such as
simulation, content creation, and autonomous driving. 3D Gaussian Splatting
(3DGS)~\citep{3dgs} enables efficient rendering through an explicit scene
representation. Building on this representation, recent feed-forward
methods~\citep{pixelsplat,Mvsplat,gs-lrm,depthsplat} directly predict Gaussian
scenes from input images, eliminating the need for per-scene representation
optimization at inference time.

Pretrained multi-view geometry models~\citep{dust3r,mast3r,vggt,Fast3r,da3,pi3}
provide a useful starting point for constructing renderable scenes from images.
They predict depth maps and cameras for each input view. Back-projecting
these depth maps with the corresponding camera estimates into a common
coordinate frame yields a predicted point cloud. This point cloud captures coherent scene geometry overall, but typically also
contains spatially isolated clusters or clusters that are inconsistent across
views. Such issues become particularly
apparent after 3D aggregation, as illustrated in Fig.~\ref{fig:teaser}.

Existing feed-forward methods~\citep{da3,anysplat,worldmirror,worldmirror-2}
combine geometry prediction from multiple views with Gaussian attribute
prediction to construct renderable scenes from input images.
For example, DA3-GS~\citep{da3} learns Gaussian attributes on top of the geometry predicted by DA3. However, as shown in
Figs.~\ref{fig:qualitative_objects} and~\ref{fig:qualitative_scenes}, renderings from DA3-GS and the other
evaluated methods still exhibit blur, ghosting, and missing detail.
Our work focuses on using these imperfect geometry predictions to construct
Gaussian scenes that support high-quality novel view rendering.

As shown in Fig.~\ref{fig:teaser}, InfiniSplat v2 reconstructs Gaussian scenes from predicted multi-view geometry and supports high-quality novel-view rendering at extreme viewpoints. We identify key factors that affect Gaussian learning and introduce three corresponding architectural and training improvements. First, we investigate why prior methods struggle to eliminate flying-point artifacts and find that jointly optimizing geometry and Gaussian attributes leads to geometric degradation. We therefore redesign the architecture to decouple geometry prediction from Gaussian attribute prediction while keeping the geometry branch frozen. Second, we find that using predicted camera poses during training~\citep{da3} introduces pose errors. Under large-angle supervision, these errors can misguide the network during training. We therefore use GT camera poses to align the predicted Gaussians. Third, inspired by InfiniSplat's strong performance under large viewpoint changes~\citep{infinisplat}, we introduce large-angle target view supervision for feed-forward Gaussian reconstruction from arbitrary input views. Our experiments show that these three improvements enhance novel view rendering quality under large viewpoint changes, even when the predicted point clouds contain numerous outliers.

We evaluate the same trained model on six datasets, both with and without camera pose inputs. 
Our method achieves the lowest mean LPIPS and DISTS across the six datasets among the evaluated baselines in
both settings. With camera pose inputs, our feed-forward reconstruction reduces LPIPS and DISTS by 25.8\% and
32.9\%, respectively, relative to the strongest evaluated baseline for each metric.
Additionally, we find that, with only image reconstruction supervision,
the trainable attribute branch learns to assign lower opacity to outliers
that are geometrically inconsistent across views. These points consequently
contribute less to the rendered images, reducing visible~artifacts~in~novel~views~(Fig.~\ref{fig:opacity_diagnostic}).

Our contributions are threefold:
\begin{itemize}
    \item We introduce InfiniSplat v2, a feed-forward Gaussian reconstruction method for arbitrary input views. It freezes the geometry branch to decouple geometry prediction from Gaussian attribute prediction, aligns predicted Gaussians using GT camera poses during training, and adopts large-angle target view supervision.

    \item We evaluate the method on six datasets in both pose-conditioned and pose-free settings. The feed-forward
    outputs of the same trained model achieve the lowest mean LPIPS and DISTS among the evaluated baselines in
    both settings.

    \item We find that opacity learned with only image reconstruction supervision
    selectively suppresses the rendering contributions of geometrically
    inconsistent points across views.
\end{itemize}

\FloatBarrier
\begin{figure}[!t]
    \centering
    \includegraphics[width=\linewidth]{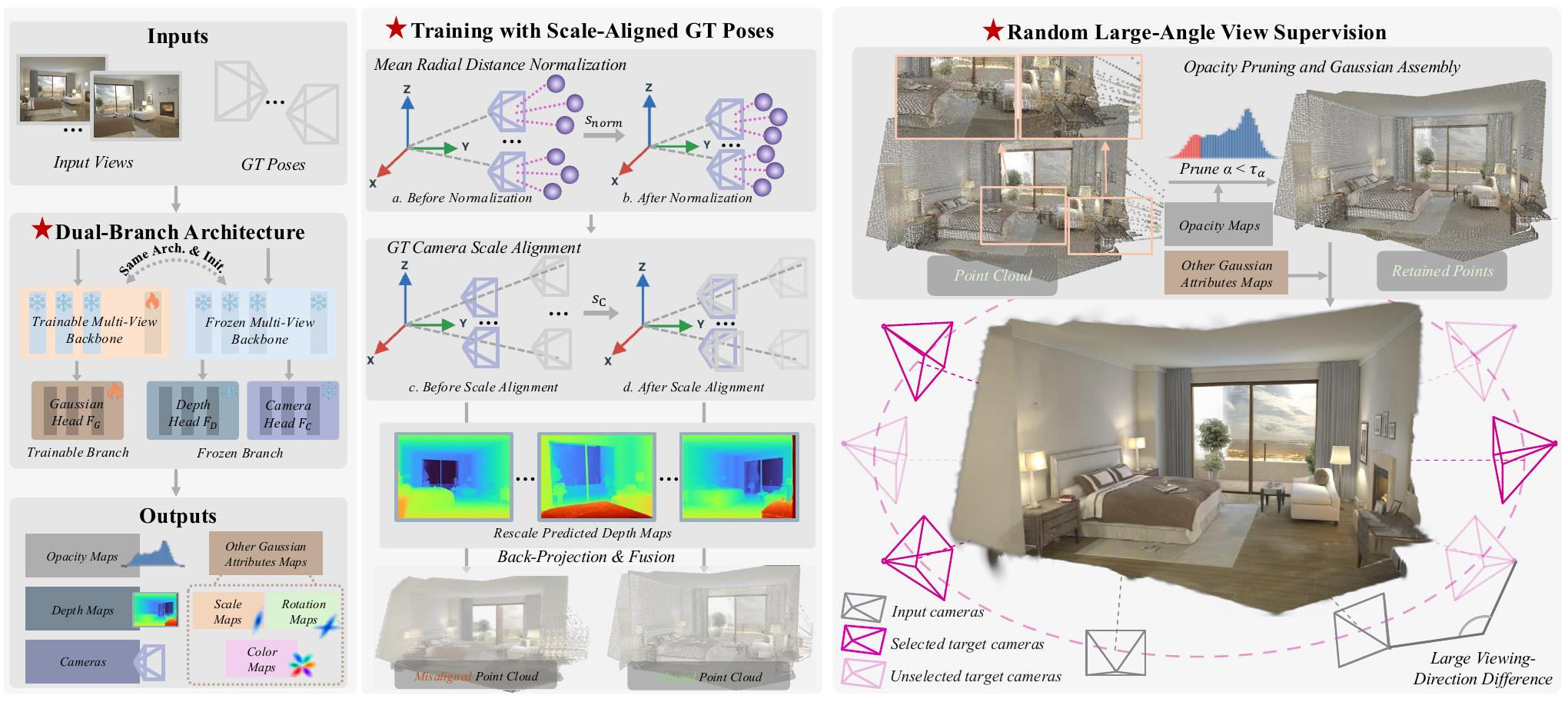}
\caption{Training pipeline of InfiniSplat v2's feed-forward reconstructor.
A frozen geometry branch predicts depth and cameras, while a trainable branch
predicts Gaussian attributes. Scale-aligned GT cameras construct the point cloud;
after opacity pruning, input and large-angle target views supervise Gaussian learning.}
    \label{fig:stage1_pipeline}
\end{figure}

\FloatBarrier
\section{Related Work}
\label{sec:related-work}

\subsection{Feed-Forward 3D Geometry Estimation}
\label{sec:rw_geometry}

Geometry foundation models recover scene structure from images and provide
a geometric basis for feed-forward Gaussian reconstruction.
DUSt3R~\citep{dust3r}, MASt3R~\citep{mast3r}, VGGT~\citep{vggt},
Fast3R~\citep{Fast3r}, and $\pi^3$~\citep{pi3} advance learned geometric
reconstruction.
Depth Anything~3 (DA3)~\citep{da3} reconstructs geometry from varying numbers
of images, with or without supplied poses. VGGT-Long~\citep{vggt-long},
VGG-T$^3$~\citep{vgg-t3}, and Scal3R~\citep{scal3r} extend learned
reconstruction to longer sequences and larger scenes. These predictions
provide a useful starting point, but constructing renderable Gaussian scenes
also requires learning attributes such as opacity, scale, and appearance.

\subsection{Feed-Forward Gaussian Splatting}
\label{sec:rw_gaussian}

Generalizable view synthesis learns to predict novel views or renderable
representations across scenes~\citep{ibrnet,lrm,lvsm}. Feed-forward Gaussian
methods pursue this goal by directly predicting explicit Gaussian scenes.
SHARP~\citep{sharp} and InfiniSplat~\citep{infinisplat} reconstruct scenes
from a single image, while GS-LRM~\citep{gs-lrm} combines multiple views.
NoPoSplat~\citep{nopoSplat} and FLARE~\citep{FLARE}
support reconstruction without supplied camera poses. Long-LRM~\citep{long-lrm}
and tttLRM~\citep{tttlrm} further extend reconstruction to long input sequences.
Work on Gaussian representations also explores different ways to organize
and predict primitives: C3G~\citep{C3G} and TokenGS~\citep{tokengs} use learnable
queries or tokens, while VolSplat~\citep{Volsplat} predicts voxel-aligned
Gaussians.

More directly related to our work is how geometry is estimated and used
for Gaussian prediction. pixelSplat~\citep{pixelsplat} establishes cross-view
correspondences through epipolar attention, while MVSplat~\citep{Mvsplat}
estimates depth using multi-view cost volumes. DepthSplat~\citep{depthsplat}
augments multi-view depth estimation with pretrained monocular depth features.
Learned geometric priors also support Gaussian prediction in
AnySplat~\citep{anysplat}, WorldMirror~\citep{worldmirror,worldmirror-2},
and DA3-GS~\citep{da3}.
Building on such priors,
SplatAnything~\citep{SplatAnything} scales up training with diverse synthetic
and real-world data. QuerySplat~\citep{QuerySplat} separately decodes
geometry and appearance from frozen geometric features.
Our work focuses on learning Gaussian attributes from imperfect predicted
point clouds to support high-quality novel view rendering.

After the initial prediction, learned refinement uses rendering feedback
to update the Gaussian scene.
G3R~\citep{g3r} and the refinement module of Flux4D~\citep{flux4d}
use explicit rendering gradients to guide learned updates.
ReSplat~\citep{resplat} instead uses feature-space rendering errors
without explicit gradient feedback. Our refiner uses input-view rendering
gradients to learn updates to Gaussian attributes while keeping centers fixed.

\section{Method}
\label{sec:method}

\paragraph{Overview.}
\label{sec:method_overview}

Given input images and optional camera conditions, InfiniSplat v2 constructs
a point cloud from predicted depth maps and corresponding cameras, then
learns Gaussian attributes for its points to regulate their rendering contributions. A frozen
DA3 branch provides the depth and camera estimates, while a trainable attribute
branch adapts to rendering
through image reconstruction supervision
(Fig.~\ref{fig:stage1_pipeline}). We then refine the scene attributes
using input-view rendering feedback.

\subsection{Background: DA3-Based Gaussian Reconstruction}
\label{sec:method_preliminaries}
Given $N$ input RGB images
$\mathcal I_{\mathcal C}=\{\mathbf I_i\}_{i\in\mathcal C}$ and optional pose
conditions $\mathbf Q_{\mathcal C}$, our goal is to predict a Gaussian scene
for novel view rendering.
Here $\mathcal C$ indexes the input views. In the pose-free setting,
our model receives only RGB images, with no supplied camera intrinsics or
poses, and $\mathbf Q_{\mathcal C}=\emptyset$.
We represent the scene as $M$ Gaussians, separating each center
$\boldsymbol\mu_m$ from its remaining attributes $\boldsymbol\xi_m$:
\begin{equation}
    \mathcal G=\{(\boldsymbol\mu_m,\boldsymbol\xi_m)\}_{m=1}^{M},
    \qquad
    \boldsymbol\xi_m=(\alpha_m,\mathbf s_m,\mathbf q_m,\mathbf c_m).
    \label{eq:gaussian_state}
\end{equation}
Here $\boldsymbol\mu_m\in\mathbb R^3$ specifies position, while
$\boldsymbol\xi_m$ contains opacity
$\alpha_m$, three-axis scale $\mathbf s_m$, rotation $\mathbf q_m$, and
zeroth-order spherical-harmonic (SH) color $\mathbf c_m$.
Scale and rotation determine the covariance and thus the Gaussian's shape.
For camera $\boldsymbol\Pi=(\mathbf K,\mathbf P)$ with intrinsics $\mathbf K$
and camera-to-world pose $\mathbf P$, the Gaussian renderer~\citep{3dgs}
produces $\widehat{\mathbf I}=\mathcal R(\mathcal G;\boldsymbol\Pi)$.

DA3-GS builds on Depth Anything~3 (DA3)~\citep{da3}, using shared multi-view
features for depth, camera, and per-pixel Gaussian prediction. Its
DPT-style Gaussian head~\citep{DPT} is trained on a frozen backbone, and an
adapter combines predicted geometry and attributes into a renderable scene.

\subsection{Feed-Forward Gaussian Reconstruction}
\label{sec:geometry_reconstructor}
Our goal is to improve Gaussian attributes through image reconstruction
supervision without compensating for attribute errors by moving point
positions. We therefore freeze the geometry branch and keep Gaussian centers
fixed at the back-projected point positions.
Constructing points with predicted input poses while supervising
them from GT target poses can also introduce camera-induced residuals;
we use scale-aligned GT cameras for both operations during training.
Finally, targets near the input viewing directions may leave Gaussian shape
insufficiently constrained. We favor large-angle target views to constrain
scales and orientations through projections from different directions.
Figure~\ref{fig:stage1_pipeline} illustrates the resulting training pipeline.

\paragraph{(1) Dual-branch architecture for attribute learning.}
To preserve DA3's depth and camera estimates while adapting Gaussian
attributes for rendering, we use two independently parameterized branches
initialized from the same pretrained DA3 weights. Given the same input images
and optional pose conditions, the frozen geometry branch predicts depths and
cameras, whereas the attribute branch produces multi-layer patch features:
\begin{equation}
    \begin{aligned}
    (\widehat D_{\mathcal C},\widehat{\boldsymbol\Pi}_{\mathcal C})
    &=\Phi_{\mathrm{geo}}(\mathcal I_{\mathcal C};\mathbf Q_{\mathcal C}),\\
    \mathbf F_{\mathcal C}
    &=\Phi_{\mathrm{attr}}(\mathcal I_{\mathcal C};\mathbf Q_{\mathcal C}).
    \end{aligned}
    \label{eq:dual_branch_prediction}
\end{equation}
Only the last $K$ Transformer blocks of $\Phi_{\mathrm{attr}}$ are fine-tuned;
$\Phi_{\mathrm{geo}}$, including its depth and camera heads, remains frozen.
A DPT Gaussian decoder maps $\mathbf F_{\mathcal C}$ to per-pixel attributes
$\boldsymbol\xi$ (Eq.~\ref{eq:gaussian_state}). The points are obtained
by the back-projection described below, without predicted positional offsets;
scales and rotations remain learnable to adjust Gaussian shape.

\paragraph{(2) Scale-aligned training.}
Errors in predicted back-projection poses introduce misalignment with
GT supervision cameras and add camera-induced residuals to the
attribute-learning objective.
We therefore use GT cameras for training-time back-projection and align
their translation scale with the predicted depths. We first express
predicted and GT poses relative to their respective reference cameras,
using the same input view as reference. Let $\mathbf x_j$ be a point back-projected using predicted depth,
intrinsics, and pose, with $\mathcal V_{\mathrm{pts}}$ indexing all input
pixels. Mean Radial Distance Normalization sets the reconstruction scale:
\begin{equation}
    s_{\mathrm{norm}}=\frac{1}{|\mathcal V_{\mathrm{pts}}|}
    \sum_{j\in\mathcal V_{\mathrm{pts}}}\|\mathbf x_j\|_2,
    \qquad \bar D_i=\frac{\widehat D_i}{s_{\mathrm{norm}}}.
    \label{eq:mean_distance_normalization}
\end{equation}
Both predicted depths and camera translations are divided by
$s_{\mathrm{norm}}$, preserving their relative scale.
To bring GT cameras to this scale, we fit a similarity transform from GT
input-camera centers to the corresponding normalized predicted camera
centers using Umeyama~\citep{Umeyama}. We retain its scale
$s_{\mathcal C}$ and preserve GT intrinsics and reference-relative rotations;
GT translations become
$\widetilde{\mathbf t}_v^\star=s_{\mathcal C}\mathbf t_v^\star$.
The same scale is applied to input and target cameras, with only input
cameras contributing to the fit.
Thus, $s_{\mathrm{norm}}$ defines the unit of the predicted reconstruction,
while $s_{\mathcal C}$ expresses GT camera translations in that unit.

Here stars denote GT quantities and tildes denote scale alignment, with
$\widetilde{\mathbf R}_i^\star=\mathbf R_i^\star$.
For each input pixel $\mathbf u=(u_x,u_y,1)^\top$, we back-project the
normalized predicted z-depth $\bar D_i$ using the scale-aligned GT camera:
\begin{equation}
    \boldsymbol\mu_{i,\mathbf u}^{\mathrm{cand}}
    =\bar D_i(\mathbf u)\,\widetilde{\mathbf R}_i^\star
    (\mathbf K_i^\star)^{-1}\mathbf u+\widetilde{\mathbf t}_i^\star.
    \label{eq:center_construction}
\end{equation}
The intrinsics $\mathbf K_i^\star$ account for image preprocessing.
Using scale-aligned GT cameras for both input-view back-projection and
target-view rendering preserves their GT relative poses within a common
reconstruction frame. This avoids the mismatch caused by constructing
points with predicted input poses while supervising them from GT
target poses.

We pair these centers with predicted per-pixel attributes and merge redundant
points into $\mathcal G_{\mathrm{vox}}$ using the voxel-based strategy of
AnySplat~\citep{anysplat} and WorldMirror~2.0~\citep{worldmirror-2}.
\paragraph{(3) Random large-angle view supervision.}
Before applying rendering supervision, we prune Gaussians with opacity below
$\tau_\alpha$ from $\mathcal G_{\mathrm{vox}}$, yielding the initial scene
$\mathcal G^0=\{(\boldsymbol\mu_m^0,\boldsymbol\xi_m^0)\}_{m=1}^{M}$,
where $M$ is the retained count. The same merging and pruning are used
at inference.

Supervision concentrated near the input viewing directions may leave
Gaussian scales and orientations insufficiently constrained. To guide
shape learning through projections across views, we favor target cameras
whose optical axes differ more from those of the input cameras. For a candidate target $v\notin\mathcal C$, let
$\mathbf f_v=\mathbf R_v^\star(0,0,1)^\top$ denote its unit optical axis
in the common reference frame. We score its angular separation from the
nearest input direction by
\begin{equation}
    a(v,\mathcal C)=\min_{i\in\mathcal C}
    \arccos(\mathbf f_v^\top\mathbf f_i).
    \label{eq:view_angle_score}
\end{equation}
We sample the target set $\mathcal T$ with a bias toward larger scores,
which measure camera orientation differences.

We supervise renderings of $\mathcal G^0$ from both input views and the
sampled target set $\mathcal T$. For target views, GT-depth warp masks
restrict the loss to valid regions with depth-consistent support from at
least one input view. Target images, cameras, and GT depth serve only
sampling and supervision; scene construction and scale estimation use
input views alone.

\subsection{Learned Rendering-Feedback Refinement}
\label{sec:fixed_center_refiner}
Starting from $\mathcal G^0$, a learned refiner updates Gaussian attributes
using input-view rendering feedback. Let $\mathcal G^t$ denote the scene
after $t$ updates. At each step, we render it into all input views,
$\mathcal F=\mathcal C$, and compute
\begin{equation}
    \begin{aligned}
    \mathbf g^t&=\operatorname{RenderGrad}
    (\mathcal G^t;\mathcal I_{\mathcal F},\boldsymbol\Pi_{\mathcal F}^{\mathrm{fb}}),\\
    \mathbf h^t&=E(\mathcal G^t,\operatorname{sg}(\mathbf g^t),\mathbf f).
    \end{aligned}
    \label{eq:rendering_feedback}
\end{equation}
Here $\boldsymbol\Pi_{\mathcal F}^{\mathrm{fb}}$ are the input cameras used
for feedback. $\operatorname{RenderGrad}$ aggregates reconstruction-loss
gradients with respect to Gaussian attributes across these views, and
$\operatorname{sg}$ denotes stop-gradient. The encoder $E$ combines this
feedback with the current scene and projected DPT features $\mathbf f$
to form per-Gaussian features $\mathbf h^t=\{\mathbf h_m^t\}_{m=1}^M$.
Projected DPT features are merged using the same voxel groups and weights
as the Gaussians to obtain $\mathbf f$, which also initializes the recurrent
hidden state.
The DPT features provide image-derived context and remain fixed across
steps, while gradient feedback is recomputed from the current scene.
Feedback uses only input observations during both training and inference.

A shared recurrent KNN Transformer $U$ aggregates features across spatial
neighbors and predicts attribute residuals and the next hidden state:
\begin{equation}
    \begin{aligned}
    (\Delta\boldsymbol\xi_m^t,\mathbf b_m^{t+1})
    &=U(\boldsymbol\xi_m^t,\mathbf h_m^t,\mathbf b_m^t),\\
    \boldsymbol\xi_m^{t+1}&=\operatorname{Apply}(\boldsymbol\xi_m^t,\Delta\boldsymbol\xi_m^t),
    \qquad \boldsymbol\mu_m^{t+1}=\boldsymbol\mu_m^0.
    \end{aligned}
    \label{eq:fixed_center_update}
\end{equation}
Here $\mathbf b_m^t$ is the recurrent hidden state, and
$\operatorname{Apply}$ applies residuals in each attribute's parameterization
space. Gaussian centers and counts remain fixed. Rendering gradients serve
as inputs to learned updates rather than direct gradient-descent steps.
The fixed centers allow the spatial KNN graph to be reused across steps.
The final scene is denoted $\mathcal G^{\mathrm{ref}}$.

\subsection{Training and Inference}
\label{sec:training_inference}
\paragraph{Training.}
In Stage~I, both branches receive input RGB and GT pose conditions.
We supervise input and sampled target renderings with equally weighted
masked $L_1$ and LPIPS~\citep{lpips} losses, updating only the attribute
branch's last $K$ Transformer blocks and Gaussian decoder. Stage~II freezes
the reconstructor and trains the feature projection and refiner.
Input-view losses provide feedback, while input and disjoint target
views supervise the updated scene.

\paragraph{Inference.}
\label{sec:inference_protocol}
The same trained model supports pose-conditioned and pose-free inference.
In the pose-conditioned setting, we construct centers using the supplied intrinsics
and poses, rescaling their translations to match the normalized predicted
depths. In the pose-free setting, we use the geometry branch's predicted
cameras, normalizing their translations by $s_{\mathrm{norm}}$.
In either setting, these cameras also render the input views for refinement
feedback. The refiner updates scene attributes while network weights remain
fixed.

The appendix specifies network architectures, gradient handling, camera
conventions, sampling, losses, and training hyperparameters.

\raggedbottom
\section{Experiments}
\label{sec:experiments}

\subsection{Experimental Setup}
\label{sec:exp_setup}

\paragraph{Datasets and baselines.}
We train InfiniSplat v2 on a mixture of Hypersim~\citep{hypersim}, Raw3D~\citep{raw3d},
and ScanNet++~\citep{scannet++}, using disjoint scenes for training and
validation. We evaluate on six datasets: DL3DV~\citep{dl3dv},
ETH3D~\citep{eth3d}, NRGBD~\citep{nrgbd}, Mip-NeRF~360~\citep{mip-nerf},
RealEstate10K (RE10K)~\citep{re10k}, and DTU~\citep{DTU}.
We compare with tttLRM~\citep{tttlrm},
AnySplat~\citep{anysplat}, C3G~\citep{C3G}, DA3-GS~\citep{da3}, DepthSplat~\citep{depthsplat},
QuerySplat~\citep{QuerySplat}, ReSplat~\citep{resplat},
TokenGS~\citep{tokengs}, VolSplat~\citep{Volsplat}, and
WorldMirror~2.0~\citep{worldmirror-2}.

\paragraph{Evaluation protocol.}
We use $N\in\{4,8,16\}$ input views and four held-out target views per
example, with the same view splits for all methods.
We evaluate methods separately in pose-conditioned and pose-free settings.
We use the same trained model for both settings. Ours denotes the
feed-forward output, and Ours$^{*}$ includes three learned refinement steps.
Evaluation-camera alignment is separate from reconstruction and uses only
input-camera correspondences.
Following SHARP~\citep{sharp}, we use LPIPS~\citep{lpips} and
DISTS~\citep{DISTS} as our primary metrics for perceptual quality.
Within each dataset, scores are computed on the same target views and valid
regions. Per-dataset scores average equally over the
three input-view counts, and the overall mean weights the six datasets equally.
The complete camera and scoring protocol is provided in the appendix.

\subsection{Comparison with State-of-the-Art Methods}
\label{sec:main_nvs}
\begin{table}[!t]
\centering
\caption{Novel view synthesis on six datasets. Scores average $N=4,8,16$ equally; Mean weights datasets equally. Ours/Ours$^{*}$: feed-forward/refined. \colorbox{nvsfirst}{First}, \colorbox{nvssecond}{second}, and \colorbox{nvsthird}{third} are highlighted per input setting; best in bold.}
\label{tab:main_nvs}
\vspace{2pt}
\begingroup
\fontsize{7}{9}\selectfont
\setlength{\tabcolsep}{1.1pt}
\renewcommand{\arraystretch}{1.12}
\resizebox{\linewidth}{!}{%
\begin{tabular}{@{}c@{\hspace{4pt}}l*{14}{c}@{}}
\toprule
\textbf{Input} & \textbf{Method} & \multicolumn{2}{c}{\textbf{DL3DV}} & \multicolumn{2}{c}{\textbf{ETH3D}} & \multicolumn{2}{c}{\textbf{NRGBD}} & \multicolumn{2}{c}{\textbf{Mip-NeRF 360}} & \multicolumn{2}{c}{\textbf{RE10K}} & \multicolumn{2}{c}{\textbf{DTU}} & \multicolumn{2}{c}{\textbf{Mean}} \\
\cmidrule(lr){3-4} \cmidrule(lr){5-6} \cmidrule(lr){7-8} \cmidrule(lr){9-10} \cmidrule(lr){11-12} \cmidrule(lr){13-14} \cmidrule(lr){15-16}
 & & LPIPS$\downarrow$ & DISTS$\downarrow$ & LPIPS$\downarrow$ & DISTS$\downarrow$ & LPIPS$\downarrow$ & DISTS$\downarrow$ & LPIPS$\downarrow$ & DISTS$\downarrow$ & LPIPS$\downarrow$ & DISTS$\downarrow$ & LPIPS$\downarrow$ & DISTS$\downarrow$ & LPIPS$\downarrow$ & DISTS$\downarrow$ \\
\midrule
\multirow{7}{*}{\rotatebox{90}{\textit{Pose-free}}} & AnySplat & 0.5985 & 0.2490 & 0.5473 & \cellcolor{nvsthird}0.2297 & \cellcolor{nvsthird}0.1895 & \cellcolor{nvsthird}0.1131 & 0.5512 & 0.2259 & 0.2454 & \cellcolor{nvsthird}0.1113 & 0.4800 & 0.2345 & 0.4353 & 0.1939 \\
 & C3G & 0.7503 & 0.3694 & 0.6722 & 0.3471 & 0.5393 & 0.2708 & 0.7390 & 0.3878 & 0.4600 & 0.2145 & 0.7115 & 0.3704 & 0.6454 & 0.3267 \\
 & QuerySplat & 0.5601 & 0.2260 & 0.5805 & 0.2750 & 0.2586 & 0.1405 & 0.6076 & 0.2678 & \cellcolor{nvsthird}0.2158 & 0.1113 & 0.5199 & 0.2722 & 0.4571 & 0.2155 \\
 & DA3-GS & 0.5146 & 0.2179 & 0.4950 & 0.2435 & 0.3456 & 0.1777 & 0.4738 & 0.2303 & 0.2526 & 0.1434 & 0.4108 & 0.2336 & 0.4154 & 0.2077 \\
 & WorldMirror 2.0 & \cellcolor{nvsthird}0.4800 & \cellcolor{nvsthird}0.2148 & \cellcolor{nvsthird}0.4917 & 0.2310 & 0.2099 & 0.1556 & \cellcolor{nvsthird}0.4701 & \cellcolor{nvsthird}0.2169 & 0.2210 & 0.1323 & \cellcolor{nvsfirst}\textbf{0.3812} & \cellcolor{nvsthird}0.2106 & \cellcolor{nvsthird}0.3757 & \cellcolor{nvsthird}0.1935 \\
 & \textbf{Ours} & \cellcolor{nvssecond}0.3773 & \cellcolor{nvsfirst}\textbf{0.1293} & \cellcolor{nvssecond}0.4145 & \cellcolor{nvsfirst}\textbf{0.1710} & \cellcolor{nvssecond}0.1096 & \cellcolor{nvssecond}0.0788 & \cellcolor{nvsfirst}\textbf{0.3784} & \cellcolor{nvsfirst}\textbf{0.1504} & \cellcolor{nvssecond}0.1459 & \cellcolor{nvssecond}0.0819 & \cellcolor{nvsthird}0.3898 & \cellcolor{nvsfirst}\textbf{0.1932} & \cellcolor{nvssecond}0.3026 & \cellcolor{nvsfirst}\textbf{0.1341} \\
 & \textbf{Ours}$^{*}$ & \cellcolor{nvsfirst}\textbf{0.3698} & \cellcolor{nvssecond}0.1306 & \cellcolor{nvsfirst}\textbf{0.4020} & \cellcolor{nvssecond}0.1738 & \cellcolor{nvsfirst}\textbf{0.1033} & \cellcolor{nvsfirst}\textbf{0.0776} & \cellcolor{nvssecond}0.3788 & \cellcolor{nvssecond}0.1595 & \cellcolor{nvsfirst}\textbf{0.1445} & \cellcolor{nvsfirst}\textbf{0.0808} & \cellcolor{nvssecond}0.3835 & \cellcolor{nvssecond}0.2048 & \cellcolor{nvsfirst}\textbf{0.2970} & \cellcolor{nvssecond}0.1379 \\
\midrule
\multirow{9}{*}{\rotatebox{90}{\textit{Pose-conditioned}}} & tttLRM & 0.4955 & 0.2574 & \cellcolor{nvsthird}0.4984 & 0.2296 & 0.2857 & 0.1979 & 0.3835 & 0.2036 & 0.2193 & 0.1404 & 0.3933 & 0.2662 & 0.3793 & 0.2159 \\
 & DA3-GS & 0.5235 & 0.2138 & 0.5776 & 0.2443 & 0.3593 & 0.1557 & 0.4965 & 0.2390 & 0.2315 & 0.1263 & 0.3536 & 0.2000 & 0.4237 & 0.1965 \\
 & DepthSplat & 0.5766 & 0.2707 & 0.5615 & 0.2874 & \cellcolor{nvsthird}0.0867 & \cellcolor{nvsthird}0.0673 & 0.4504 & 0.2448 & 0.1297 & \cellcolor{nvsthird}0.0840 & 0.4263 & 0.2413 & 0.3718 & 0.1993 \\
 & ReSplat & \cellcolor{nvsthird}0.4198 & \cellcolor{nvsthird}0.2123 & 0.5266 & 0.2309 & 0.1204 & 0.0962 & \cellcolor{nvsthird}0.3655 & \cellcolor{nvsthird}0.1747 & \cellcolor{nvssecond}0.1191 & 0.0842 & 0.4485 & 0.2781 & \cellcolor{nvsthird}0.3333 & 0.1794 \\
 & TokenGS & 0.6911 & 0.3239 & 0.6704 & 0.2913 & 0.5589 & 0.3025 & 0.6456 & 0.3095 & 0.3955 & 0.2094 & 0.5057 & 0.3221 & 0.5779 & 0.2931 \\
 & VolSplat & 0.7049 & 0.3539 & 0.6715 & 0.3149 & 0.2342 & 0.1911 & 0.5071 & 0.2585 & 0.3552 & 0.1985 & 0.6424 & 0.3370 & 0.5192 & 0.2756 \\
 & WorldMirror 2.0 & 0.4673 & 0.2126 & 0.5201 & \cellcolor{nvsthird}0.2079 & 0.1056 & 0.0717 & 0.4689 & 0.2194 & 0.2174 & 0.1359 & \cellcolor{nvsthird}0.3306 & \cellcolor{nvsthird}0.1930 & 0.3517 & \cellcolor{nvsthird}0.1734 \\
 & \textbf{Ours} & \cellcolor{nvssecond}0.3363 & \cellcolor{nvssecond}0.1209 & \cellcolor{nvssecond}0.4339 & \cellcolor{nvssecond}0.1661 & \cellcolor{nvssecond}0.0653 & \cellcolor{nvssecond}0.0535 & \cellcolor{nvssecond}0.3206 & \cellcolor{nvssecond}0.1405 & \cellcolor{nvsthird}0.1202 & \cellcolor{nvsfirst}\textbf{0.0798} & \cellcolor{nvssecond}0.2070 & \cellcolor{nvsfirst}\textbf{0.1368} & \cellcolor{nvssecond}0.2472 & \cellcolor{nvssecond}0.1163 \\
 & \textbf{Ours}$^{*}$ & \cellcolor{nvsfirst}\textbf{0.3276} & \cellcolor{nvsfirst}\textbf{0.1197} & \cellcolor{nvsfirst}\textbf{0.4176} & \cellcolor{nvsfirst}\textbf{0.1631} & \cellcolor{nvsfirst}\textbf{0.0612} & \cellcolor{nvsfirst}\textbf{0.0524} & \cellcolor{nvsfirst}\textbf{0.3099} & \cellcolor{nvsfirst}\textbf{0.1380} & \cellcolor{nvsfirst}\textbf{0.1176} & \cellcolor{nvssecond}0.0802 & \cellcolor{nvsfirst}\textbf{0.2014} & \cellcolor{nvssecond}0.1422 & \cellcolor{nvsfirst}\textbf{0.2392} & \cellcolor{nvsfirst}\textbf{0.1159} \\
\bottomrule
\end{tabular}%
}
\endgroup
\end{table}

\paragraph{Quantitative comparison.}
Across six datasets, our feed-forward model achieves the best mean LPIPS
and DISTS among evaluated baselines in both settings (Table~\ref{tab:main_nvs}).
With pose conditioning, LPIPS improves by 25.8\% over ReSplat and DISTS
by 32.9\% over WorldMirror~2.0; without poses, the gains over
WorldMirror~2.0 are 19.5\% and 30.7\%, respectively.
Results for all four metrics and a comparison of learned refinement with Adam
are provided in the appendix.

\begin{figure}[!t]
    \centering
    \includegraphics[width=\linewidth]{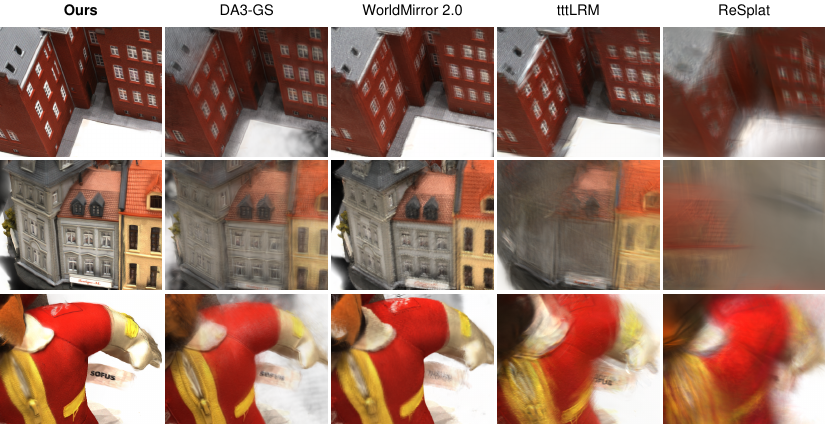}
\caption{Object-level qualitative comparisons under multi-view input settings.}
    \label{fig:qualitative_comparison}
    \label{fig:qualitative_objects}
\end{figure}

\begin{figure}[!t]
    \centering
    \includegraphics[width=\linewidth]{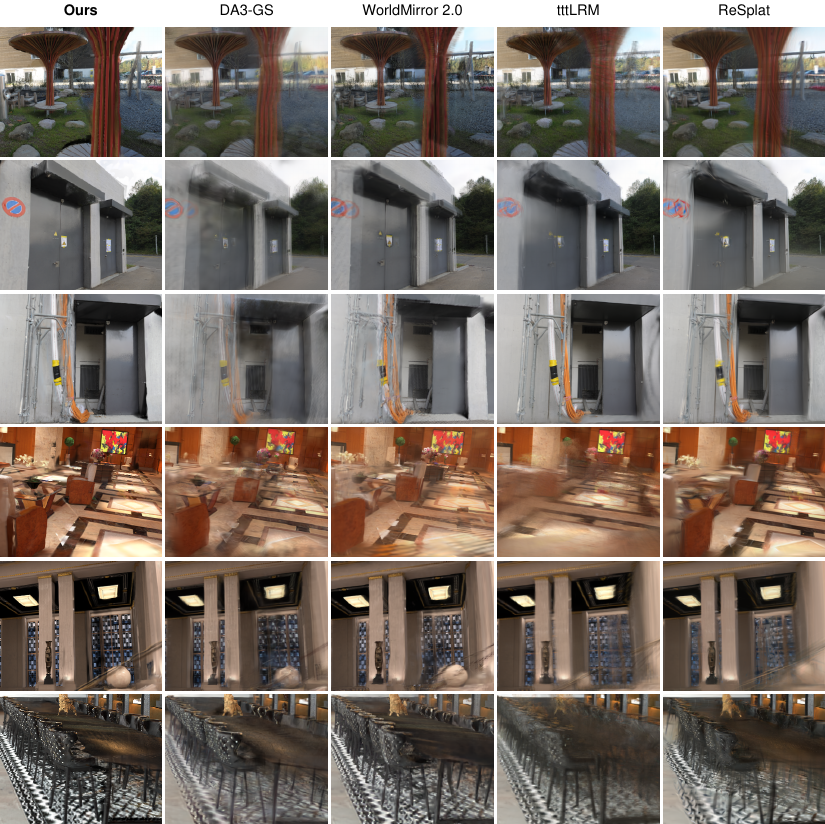}

\caption{Scene-level qualitative comparisons under multi-view input settings.}
    \label{fig:qualitative_scenes}
\end{figure}

\paragraph{Qualitative evaluation.}
Figures~\ref{fig:qualitative_objects} and~\ref{fig:qualitative_scenes}
show that our method preserves fine textures and structural details in both
object-level and scene-level reconstructions. More qualitative results are
provided in the appendix.

\subsection{Ablations and Analysis}
\label{sec:exp_ablation}
We evaluate attribute-backbone fine-tuning (A), geometry freezing (G),
GT-camera backprojection (C), and large-angle supervision (V) on ETH3D
and DTU, averaging pre-refinement results over $N=4,8,16$ input views.
We additionally evaluate learned positional offsets (w/ O).
Ours uses $K=10$; w/o A uses $K=0$.

\begin{figure}[!t]
\centering
\begin{minipage}[t]{0.49\linewidth}
\vspace{0pt}
\centering
\begingroup
\small
\setlength{\tabcolsep}{2pt}
\begin{tabular}{lcccc}
\toprule
\multirow{2}{*}{Variant} & \multicolumn{2}{c}{ETH3D} & \multicolumn{2}{c}{DTU} \\
\cmidrule(lr){2-3}\cmidrule(lr){4-5}
& LPIPS$\downarrow$ & DISTS$\downarrow$ & LPIPS$\downarrow$ & DISTS$\downarrow$ \\
\midrule
Ours & \cellcolor{nvsfirst}\textbf{0.4339} & \cellcolor{nvsfirst}\textbf{0.1661} & \cellcolor{nvsfirst}\textbf{0.2070} & \cellcolor{nvsfirst}\textbf{0.1368} \\
\midrule
w/o A & \cellcolor{nvssecond}0.4388 & 0.2019 & \cellcolor{nvsthird}0.2612 & \cellcolor{nvsthird}0.1663 \\
w/o G & 0.5204 & 0.1938 & 0.2695 & 0.1841 \\
w/ O & 0.4998 & 0.2082 & 0.2826 & 0.1887 \\
w/o C & 0.5017 & \cellcolor{nvsthird}0.1863 & 0.3237 & 0.1758 \\
w/o V & \cellcolor{nvsthird}0.4533 & \cellcolor{nvssecond}0.1763 & \cellcolor{nvssecond}0.2295 & \cellcolor{nvssecond}0.1508 \\
\bottomrule
\end{tabular}
\endgroup
\vspace{6pt}
\makeatletter
\def\@captype{table}
\makeatother
\caption{Ablation study of reconstruction components on ETH3D and DTU.}
\label{tab:ablation_reconstructor_v2}
\end{minipage}\hfill
\begin{minipage}[t]{0.485\linewidth}
\vspace{0pt}
\centering
\input{figures/offset_kitchen_main}
\caption{Effect of positional offsets on rendering and depth in an NRGBD scene.}
\label{fig:offset_kitchen_main}
\end{minipage}
\end{figure}

Under image reconstruction supervision, geometry fine-tuning (w/o G) and
learned offsets (w/ O) worsen both metrics on ETH3D and DTU
(Table~\ref{tab:ablation_reconstructor_v2}). Freezing geometry instead of
fine-tuning reduces LPIPS by 16.6\% and 23.2\%, respectively.
We interpret this degradation as positional updates compensating for
attribute errors, a pathway blocked by fixed centers.
With the geometry backbone frozen, w/ O predicts 3D offsets to back-projected
centers through the Gaussian head, producing local blur and depth distortions
in the NRGBD kitchen example (Fig.~\ref{fig:offset_kitchen_main}).

Fine-tuning the last $K=10$ attribute-backbone blocks, compared with
$K=0$ (w/o A), reduces LPIPS by 1.1\% on ETH3D and 20.8\% on DTU,
with lower DISTS on both datasets. GT-camera backprojection reduces
DTU LPIPS by 36.1\% relative to predicted cameras (w/o C).
Large-angle target view supervision improves both LPIPS and DISTS on
ETH3D and DTU compared with local interpolation-view supervision (w/o V).

\paragraph{Qualitative ablations.}
In Fig.~\ref{fig:ablation_feedforward}, w/o A and w/o V produce pronounced
cracks on the tabletop in the upper row; w/o G, w/ O, and w/o C exhibit
milder cracks. In the lower row, w/o G and w/o C blur the door frames,
which remain more distinct in the full model.

\begingroup
\setlength{\intextsep}{4pt}
\setlength{\abovecaptionskip}{4pt}
\begin{figure}[!t]
    \centering
    \input{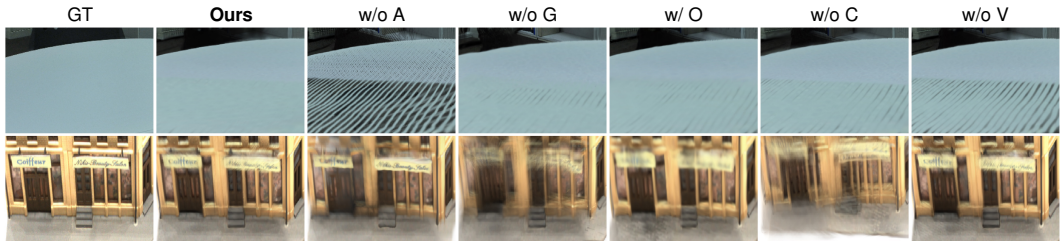}
\caption{Effect of reconstruction components on surface artifacts and structural detail.}
    \label{fig:ablation_feedforward}
\end{figure}

\paragraph{Opacity and geometric consistency.}
Using GT-depth labels on held-out Hypersim~\citep{hypersim} validation
scenes, we analyze opacity learned with only image reconstruction
supervision.
Labels are independent of opacity: inconsistent points violate GT-observed
free space in at least two views and agree with no observed surface depth.
At threshold 0.01, pruning removes 51.8\% of geometrically
inconsistent points but only 7.5\% of all points, retaining
93.3\% of surface-consistent points. This selective removal links
opacity to geometric reliability while preserving the main scene
structure and appearance (Fig.~\ref{fig:opacity_diagnostic}).

\begin{figure}[!t]
\centering
\includegraphics[width=\linewidth]{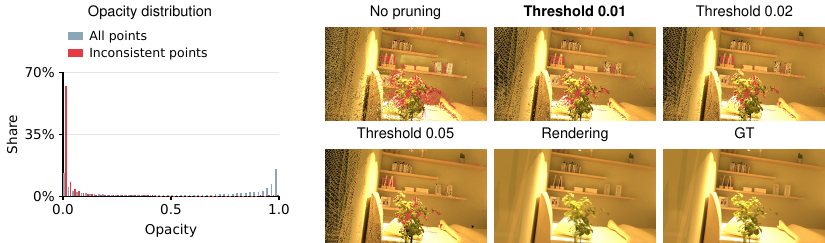}
\caption{Opacity distributions and the effect of pruning on geometrically inconsistent points.
GT depth independently defines the inconsistent subset; distributions are normalized separately.}
\label{fig:opacity_diagnostic}
\end{figure}
\endgroup

\FloatBarrier
\section{Conclusion}
\label{sec:conclusion}
InfiniSplat v2 improves mean LPIPS and DISTS across six datasets in both
pose settings. ETH3D and DTU ablations support fixed centers and attribute
adaptation under reconstruction supervision. Learned opacity preferentially
suppresses geometrically inconsistent points.

\flushbottom
\subsection*{AI Use Statement}
Generative AI tools were used to assist with language polishing of
author-written text and with code development and debugging. The authors
reviewed and approved the revised text and inspected and tested the
AI-assisted code.
Generative AI tools were not used to originate the research ideas or core
technical contributions, or to generate experimental or training data.
References were independently identified, collected, and selected by the
authors. The authors take full responsibility for the final content of
this work, including its text, claims, experimental results, and code.

\bibliographystyle{iclr2027_conference}
\bibliography{references}

\begin{thebibliography}{51}
\providecommand{\natexlab}[1]{#1}
\providecommand{\url}[1]{\texttt{#1}}
\expandafter\ifx\csname urlstyle\endcsname\relax
  \providecommand{\doi}[1]{doi: #1}\else
  \providecommand{\doi}{doi: \begingroup \urlstyle{rm}\Url}\fi

\bibitem[Aan{\ae}s et~al.(2016)Aan{\ae}s, Jensen, Vogiatzis, Tola, and
  Dahl]{DTU}
Henrik Aan{\ae}s, Rasmus~Ramsb{\o}l Jensen, George Vogiatzis, Engin Tola, and
  Anders~Bjorholm Dahl.
\newblock Large-scale data for multiple-view stereopsis.
\newblock \emph{International Journal of Computer Vision}, 120\penalty0
  (2):\penalty0 153--168, 2016.

\bibitem[An et~al.(2025)An, Jung, Kim, Hong, Kim, Fukuda, Jeon, Han, Narihira,
  Ko, et~al.]{C3G}
Honggyu An, Jaewoo Jung, Mungyeom Kim, Sunghwan Hong, Chaehyun Kim, Kazumi
  Fukuda, Minkyeong Jeon, Jisang Han, Takuya Narihira, Hyuna Ko, et~al.
\newblock C3g: Learning compact 3d representations with 2k gaussians.
\newblock \emph{arXiv preprint arXiv:2512.04021}, 2025.

\bibitem[Anonymous(2027)]{SplatAnything}
Anonymous.
\newblock Splatanything: Gaussian spacing normalization for feed-forward 3d
  gaussian splatting across arbitrary scenes, 2027.
\newblock Manuscript under review at ICLR 2027.

\bibitem[Azinovi\'c et~al.(2022)Azinovi\'c, Martin-Brualla, Goldman,
  Nie{\ss}ner, and Thies]{nrgbd}
Dejan Azinovi\'c, Ricardo Martin-Brualla, Dan~B Goldman, Matthias Nie{\ss}ner,
  and Justus Thies.
\newblock Neural rgb-d surface reconstruction.
\newblock In \emph{Proceedings of the IEEE/CVF Conference on Computer Vision
  and Pattern Recognition (CVPR)}, pp.\  6290--6301, June 2022.

\bibitem[Barron et~al.(2022)Barron, Mildenhall, Verbin, Srinivasan, and
  Hedman]{mip-nerf}
Jonathan~T Barron, Ben Mildenhall, Dor Verbin, Pratul~P Srinivasan, and Peter
  Hedman.
\newblock Mip-nerf 360: Unbounded anti-aliased neural radiance fields.
\newblock In \emph{2022 IEEE/CVF Conference on Computer Vision and Pattern
  Recognition (CVPR)}, pp.\  5460--5469. IEEE, 2022.

\bibitem[Charatan et~al.(2024)Charatan, Li, Tagliasacchi, and
  Sitzmann]{pixelsplat}
David Charatan, Sizhe~Lester Li, Andrea Tagliasacchi, and Vincent Sitzmann.
\newblock pixelsplat: 3d gaussian splats from image pairs for scalable
  generalizable 3d reconstruction.
\newblock In \emph{2024 IEEE/CVF Conference on Computer Vision and Pattern
  Recognition (CVPR)}, pp.\  19457--19467. IEEE, 2024.

\bibitem[Chen et~al.(2024{\natexlab{a}})Chen, Xu, Zheng, Zhuang, Pollefeys,
  Geiger, Cham, and Cai]{Mvsplat}
Yuedong Chen, Haofei Xu, Chuanxia Zheng, Bohan Zhuang, Marc Pollefeys, Andreas
  Geiger, Tat-Jen Cham, and Jianfei Cai.
\newblock Mvsplat: Efficient 3d gaussian splatting from sparse multi-view
  images.
\newblock In \emph{European conference on computer vision}, pp.\  370--386.
  Springer, 2024{\natexlab{a}}.

\bibitem[Chen et~al.(2024{\natexlab{b}})Chen, Wang, Yang, Manivasagam, and
  Urtasun]{g3r}
Yun Chen, Jingkang Wang, Ze~Yang, Sivabalan Manivasagam, and Raquel Urtasun.
\newblock G3r: Gradient guided generalizable reconstruction.
\newblock In \emph{European Conference on Computer Vision}, pp.\  305--323.
  Springer, 2024{\natexlab{b}}.

\bibitem[Deng et~al.(2025)Deng, Ti, Xu, Yang, and Xie]{vggt-long}
Kai Deng, Zexin Ti, Jiawei Xu, Jian Yang, and Jin Xie.
\newblock Vggt-long: Chunk it, loop it, align it--pushing vggt's limits on
  kilometer-scale long rgb sequences.
\newblock \emph{arXiv preprint arXiv:2507.16443}, 2025.

\bibitem[Ding et~al.(2020)Ding, Ma, Wang, and Simoncelli]{DISTS}
Keyan Ding, Kede Ma, Shiqi Wang, and Eero~P Simoncelli.
\newblock Image quality assessment: Unifying structure and texture similarity.
\newblock \emph{IEEE transactions on pattern analysis and machine
  intelligence}, 44\penalty0 (5):\penalty0 2567--2581, 2020.

\bibitem[Elflein et~al.(2026)Elflein, Li, Agostinho, Gojcic, Leal-Taix{\'e},
  Zhou, and Osep]{vgg-t3}
Sven Elflein, Ruilong Li, S{\'e}rgio Agostinho, Zan Gojcic, Laura
  Leal-Taix{\'e}, Qunjie Zhou, and Aljosa Osep.
\newblock {VGG-T3}: Offline feed-forward 3d reconstruction at scale.
\newblock \emph{arXiv preprint arXiv:2602.23361}, 2026.

\bibitem[Geiger et~al.(2013)Geiger, Lenz, Stiller, and Urtasun]{KITTI}
Andreas Geiger, Philip Lenz, Christoph Stiller, and Raquel Urtasun.
\newblock Vision meets robotics: The kitti dataset.
\newblock \emph{International Journal of Robotics Research (IJRR)}, 2013.

\bibitem[Guizilini et~al.(2020)Guizilini, Ambrus, Pillai, Raventos, and
  Gaidon]{ddad}
Vitor Guizilini, Rares Ambrus, Sudeep Pillai, Allan Raventos, and Adrien
  Gaidon.
\newblock 3d packing for self-supervised monocular depth estimation.
\newblock In \emph{Proceedings of the IEEE/CVF conference on computer vision
  and pattern recognition}, pp.\  2485--2494, 2020.

\bibitem[Hong et~al.(2024)Hong, Zhang, Gu, Bi, Zhou, Liu, Liu, Sunkavalli, Bui,
  and Tan]{lrm}
Yicong Hong, Kai Zhang, Jiuxiang Gu, Sai Bi, Yang Zhou, Difan Liu, Feng Liu,
  Kalyan Sunkavalli, Trung Bui, and Hao Tan.
\newblock Lrm: Large reconstruction model for single image to 3d.
\newblock In \emph{International Conference on Learning Representations},
  volume 2024, pp.\  50678--50702, 2024.

\bibitem[Jiang et~al.(2025)Jiang, Mao, Xu, Lu, Ren, Jin, Xu, Yu, Pang, Zhao,
  et~al.]{anysplat}
Lihan Jiang, Yucheng Mao, Linning Xu, Tao Lu, Kerui Ren, Yichen Jin, Xudong Xu,
  Mulin Yu, Jiangmiao Pang, Feng Zhao, et~al.
\newblock Anysplat: Feed-forward 3d gaussian splatting from unconstrained
  views.
\newblock \emph{ACM Transactions on Graphics (TOG)}, 44\penalty0 (6):\penalty0
  1--16, 2025.

\bibitem[Jin et~al.(2024)Jin, Jiang, Tan, Zhang, Bi, Zhang, Luan, Snavely, and
  Xu]{lvsm}
Haian Jin, Hanwen Jiang, Hao Tan, Kai Zhang, Sai Bi, Tianyuan Zhang, Fujun
  Luan, Noah Snavely, and Zexiang Xu.
\newblock Lvsm: A large view synthesis model with minimal 3d inductive bias.
\newblock \emph{arXiv preprint arXiv:2410.17242}, 2024.

\bibitem[Kerbl et~al.(2023)Kerbl, Kopanas, Leimk{\"u}hler, Drettakis,
  et~al.]{3dgs}
Bernhard Kerbl, Georgios Kopanas, Thomas Leimk{\"u}hler, George Drettakis,
  et~al.
\newblock 3d gaussian splatting for real-time radiance field rendering.
\newblock \emph{ACM Trans. Graph.}, 42\penalty0 (4):\penalty0 139--1, 2023.

\bibitem[Leroy et~al.(2024)Leroy, Cabon, and Revaud]{mast3r}
Vincent Leroy, Yohann Cabon, and J{\'e}r{\^o}me Revaud.
\newblock Grounding image matching in 3d with mast3r.
\newblock In \emph{European conference on computer vision}, pp.\  71--91.
  Springer, 2024.

\bibitem[Li et~al.(2026)Li, Shen, Zhang, Ye, Wu, Hao, Zhang, and
  Liu]{QuerySplat}
Yinglong Li, Donghui Shen, Xiaoyu Zhang, Zhichao Ye, Hongyu Wu, Aimin Hao,
  Guofeng Zhang, and Haomin Liu.
\newblock Querysplat: Decoupling geometry and appearance representations in
  3dgs prediction.
\newblock \emph{arXiv preprint arXiv:2608.01186}, 2026.

\bibitem[Lin et~al.(2025)Lin, Chen, Liew, Chen, Li, Shi, Feng, and Kang]{da3}
Haotong Lin, Sili Chen, Junhao Liew, Donny~Y Chen, Zhenyu Li, Guang Shi, Jiashi
  Feng, and Bingyi Kang.
\newblock Depth anything 3: Recovering the visual space from any views.
\newblock \emph{arXiv preprint arXiv:2511.10647}, 2025.

\bibitem[Ling et~al.(2024)Ling, Sheng, Tu, Zhao, Xin, Wan, Yu, Guo, Yu, Lu,
  et~al.]{dl3dv}
Lu~Ling, Yichen Sheng, Zhi Tu, Wentian Zhao, Cheng Xin, Kun Wan, Lantao Yu,
  Qianyu Guo, Zixun Yu, Yawen Lu, et~al.
\newblock Dl3dv-10k: A large-scale scene dataset for deep learning-based 3d
  vision.
\newblock In \emph{2024 IEEE/CVF Conference on Computer Vision and Pattern
  Recognition (CVPR)}, pp.\  22160--22169. IEEE, 2024.

\bibitem[Liu et~al.(2025)Liu, Min, Wang, Wu, Wang, Yuan, Luo, and
  Guo]{worldmirror}
Yifan Liu, Zhiyuan Min, Zhenwei Wang, Junta Wu, Tengfei Wang, Yixuan Yuan,
  Yawei Luo, and Chunchao Guo.
\newblock Worldmirror: Universal 3d world reconstruction with any-prior
  prompting.
\newblock \emph{arXiv preprint arXiv:2510.10726}, 2025.

\bibitem[Mescheder et~al.(2026)Mescheder, Dong, Li, Bai, Santos, Hu, Lecouat,
  Zhen, Delaunoy, Fang, et~al.]{sharp}
Lars Mescheder, Wei Dong, Shiwei Li, Xuyang Bai, Marcel Santos, Peiyun Hu,
  Bruno Lecouat, Mingmin Zhen, Ama{\"e}l Delaunoy, Tian Fang, et~al.
\newblock Sharp monocular view synthesis in less than a second.
\newblock In \emph{International Conference on Learning Representations},
  volume 2026, pp.\  24192--24230, 2026.

\bibitem[Ranftl et~al.(2021)Ranftl, Bochkovskiy, and Koltun]{DPT}
Ren{\'e} Ranftl, Alexey Bochkovskiy, and Vladlen Koltun.
\newblock Vision transformers for dense prediction.
\newblock In \emph{2021 IEEE/CVF International Conference on Computer Vision
  (ICCV)}, pp.\  12159--12168. IEEE, 2021.

\bibitem[Ren et~al.(2026)Ren, Tyszkiewicz, Huang, and Gojcic]{tokengs}
Jiawei Ren, Michal~Jan Tyszkiewicz, Jiahui Huang, and Zan Gojcic.
\newblock Tokengs: Decoupling 3d gaussian prediction from pixels with learnable
  tokens.
\newblock \emph{arXiv preprint arXiv:2604.15239}, 2026.

\bibitem[Roberts et~al.(2021)Roberts, Ramapuram, Ranjan, Kumar, Bautista,
  Paczan, Webb, and Susskind]{hypersim}
Mike Roberts, Jason Ramapuram, Anurag Ranjan, Atulit Kumar, Miguel~Angel
  Bautista, Nathan Paczan, Russ Webb, and Joshua~M Susskind.
\newblock Hypersim: A photorealistic synthetic dataset for holistic indoor
  scene understanding.
\newblock In \emph{2021 IEEE/CVF International Conference on Computer Vision
  (ICCV)}, pp.\  10892--10902. IEEE, 2021.

\bibitem[Schops et~al.(2017)Schops, Schonberger, Galliani, Sattler, Schindler,
  Pollefeys, and Geiger]{eth3d}
Thomas Schops, Johannes~L Schonberger, Silvano Galliani, Torsten Sattler,
  Konrad Schindler, Marc Pollefeys, and Andreas Geiger.
\newblock A multi-view stereo benchmark with high-resolution images and
  multi-camera videos.
\newblock In \emph{Proceedings of the IEEE conference on computer vision and
  pattern recognition}, pp.\  3260--3269, 2017.

\bibitem[Sun et~al.(2020)Sun, Kretzschmar, Dotiwalla, Chouard, Patnaik, Tsui,
  Guo, Zhou, Chai, Caine, et~al.]{waymo}
Pei Sun, Henrik Kretzschmar, Xerxes Dotiwalla, Aurelien Chouard, Vijaysai
  Patnaik, Paul Tsui, James Guo, Yin Zhou, Yuning Chai, Benjamin Caine, et~al.
\newblock Scalability in perception for autonomous driving: Waymo open dataset.
\newblock In \emph{2020 IEEE/CVF Conference on Computer Vision and Pattern
  Recognition (CVPR)}, pp.\  2443--2451. IEEE, 2020.

\bibitem[Team et~al.(2026)Team, Cao, Zuo, Wang, Zhang, Wu, Liu, Gong, Liu,
  Yuan, et~al.]{worldmirror-2}
HY~Team, Chenjie Cao, Xuhui Zuo, Zhenwei Wang, Yisu Zhang, Junta Wu, Zhenyang
  Liu, Yuning Gong, Yang Liu, Bo~Yuan, et~al.
\newblock Hy-world 2.0: A multi-modal world model for reconstructing,
  generating, and simulating 3d worlds.
\newblock \emph{arXiv e-prints}, pp.\  arXiv--2604, 2026.

\bibitem[Team(2026)]{raw3d}
Rawmantic Team.
\newblock Rawmantic 3d training field: From the real world, for the real world.
\newblock \emph{Rawmantic}, 2026.

\bibitem[Turki et~al.(2022)Turki, Ramanan, and Satyanarayanan]{mega-nerf}
Haithem Turki, Deva Ramanan, and Mahadev Satyanarayanan.
\newblock Mega-nerf: Scalable construction of large-scale nerfs for virtual
  fly-throughs.
\newblock In \emph{2022 IEEE/CVF Conference on Computer Vision and Pattern
  Recognition (CVPR)}, pp.\  12912--12921. IEEE, 2022.

\bibitem[Umeyama(1991)]{Umeyama}
Shinji Umeyama.
\newblock Least-squares estimation of transformation parameters between two
  point patterns.
\newblock \emph{IEEE Transactions on pattern analysis and machine
  intelligence}, 13\penalty0 (4):\penalty0 376--380, 1991.

\bibitem[Wang et~al.(2026{\natexlab{a}})Wang, Tan, Yifan, Chen, Liu,
  Sunkavalli, Bi, Liu, and Hu]{tttlrm}
Chen Wang, Hao Tan, Wang Yifan, Zhiqin Chen, Yuheng Liu, Kalyan Sunkavalli, Sai
  Bi, Lingjie Liu, and Yiwei Hu.
\newblock tttlrm: Test-time training for long context and autoregressive 3d
  reconstruction.
\newblock \emph{arXiv preprint arXiv:2602.20160}, 2026{\natexlab{a}}.

\bibitem[Wang et~al.(2025{\natexlab{a}})Wang, Chen, Karaev, Vedaldi, Rupprecht,
  and Novotny]{vggt}
Jianyuan Wang, Minghao Chen, Nikita Karaev, Andrea Vedaldi, Christian
  Rupprecht, and David Novotny.
\newblock Vggt: Visual geometry grounded transformer.
\newblock In \emph{2025 IEEE/CVF Conference on Computer Vision and Pattern
  Recognition (CVPR)}, pp.\  5294--5306. IEEE, 2025{\natexlab{a}}.

\bibitem[Wang et~al.(2026{\natexlab{b}})Wang, Yu, Hu, Yang, Ni, Zhan, Chen,
  Zhou, Hu, and Peng]{infinisplat}
Jiawei Wang, Hao Yu, Yongzhen Hu, Xinyi Yang, Tao Ni, Xin Zhan, Junbo Chen,
  Xiaowei Zhou, Ruizhen Hu, and Sida Peng.
\newblock Infinisplat: Implicit gaussian decoding for large-baseline monocular
  view synthesis.
\newblock \emph{arXiv preprint arXiv:2608.02437}, 2026{\natexlab{b}}.

\bibitem[Wang et~al.(2026{\natexlab{c}})Wang, Che, Chen, Yang, Goli,
  Manivasagam, and Urtasun]{flux4d}
Jingkang Wang, Henry Che, Yun Chen, Ze~Yang, Lily Goli, Sivabalan Manivasagam,
  and Raquel Urtasun.
\newblock Flux4d: Flow-based unsupervised 4d reconstruction.
\newblock \emph{Advances in Neural Information Processing Systems},
  38:\penalty0 105673--105693, 2026{\natexlab{c}}.

\bibitem[Wang et~al.(2021)Wang, Wang, Genova, Srinivasan, Zhou, Barron,
  Martin-Brualla, Snavely, and Funkhouser]{ibrnet}
Qianqian Wang, Zhicheng Wang, Kyle Genova, Pratul~P Srinivasan, Howard Zhou,
  Jonathan~T Barron, Ricardo Martin-Brualla, Noah Snavely, and Thomas
  Funkhouser.
\newblock Ibrnet: Learning multi-view image-based rendering.
\newblock In \emph{Proceedings of the IEEE/CVF conference on computer vision
  and pattern recognition}, pp.\  4690--4699, 2021.

\bibitem[Wang et~al.(2024)Wang, Leroy, Cabon, Chidlovskii, and Revaud]{dust3r}
Shuzhe Wang, Vincent Leroy, Yohann Cabon, Boris Chidlovskii, and Jerome Revaud.
\newblock Dust3r: Geometric 3d vision made easy.
\newblock In \emph{2024 IEEE/CVF Conference on Computer Vision and Pattern
  Recognition (CVPR)}, pp.\  20697--20709. IEEE, 2024.

\bibitem[Wang et~al.(2025{\natexlab{b}})Wang, Chen, Zhang, Liu, Wang, Feng,
  Qin, Chen, Bian, Zhu, et~al.]{Volsplat}
Weijie Wang, Yeqing Chen, Zeyu Zhang, Hengyu Liu, Haoxiao Wang, Zhiyuan Feng,
  Wenkang Qin, Feng Chen, Jia-Wang Bian, Zheng Zhu, et~al.
\newblock Volsplat: Rethinking feed-forward 3d gaussian splatting with
  voxel-aligned prediction.
\newblock \emph{arXiv preprint arXiv:2509.19297}, 2025{\natexlab{b}}.

\bibitem[Wang et~al.(2026{\natexlab{d}})Wang, Zhou, Zhu, Chang, Zhou, Li, Chen,
  Pang, Shen, and He]{pi3}
Yifan Wang, Jianjun Zhou, Haoyi Zhu, Wenzheng Chang, Yang Zhou, Zizun Li, Junyi
  Chen, Jiangmiao Pang, Chunhua Shen, and Tong He.
\newblock {Pi3}: Permutation-equivariant visual geometry learning.
\newblock In \emph{International Conference on Learning Representations},
  volume 2026, pp.\  10481--10497, 2026{\natexlab{d}}.

\bibitem[Xie et~al.(2026)Xie, Yang, Jin, Cai, Yin, Ren, Zhang, Hua, Peng, Guo,
  et~al.]{scal3r}
Tao Xie, Peishan Yang, Yudong Jin, Yingfeng Cai, Wei Yin, Weiqiang Ren, Qian
  Zhang, Wei Hua, Sida Peng, Xiaoyang Guo, et~al.
\newblock Scal3r: Scalable test-time training for large-scale 3d
  reconstruction.
\newblock \emph{arXiv preprint arXiv:2604.08542}, 2026.

\bibitem[Xu et~al.(2025{\natexlab{a}})Xu, Barath, Geiger, and
  Pollefeys]{resplat}
Haofei Xu, Daniel Barath, Andreas Geiger, and Marc Pollefeys.
\newblock Resplat: Learning recurrent gaussian splatting.
\newblock \emph{arXiv preprint arXiv:2510.08575}, 2025{\natexlab{a}}.

\bibitem[Xu et~al.(2025{\natexlab{b}})Xu, Peng, Wang, Blum, Barath, Geiger, and
  Pollefeys]{depthsplat}
Haofei Xu, Songyou Peng, Fangjinhua Wang, Hermann Blum, Daniel Barath, Andreas
  Geiger, and Marc Pollefeys.
\newblock Depthsplat: Connecting gaussian splatting and depth.
\newblock In \emph{2025 IEEE/CVF Conference on Computer Vision and Pattern
  Recognition (CVPR)}, pp.\  16453--16463. IEEE, 2025{\natexlab{b}}.

\bibitem[Yang et~al.(2025)Yang, Sax, Liang, Henaff, Tang, Cao, Chai, Meier, and
  Feiszli]{Fast3r}
Jianing Yang, Alexander Sax, Kevin~J Liang, Mikael Henaff, Hao Tang, Ang Cao,
  Joyce Chai, Franziska Meier, and Matt Feiszli.
\newblock Fast3r: Towards 3d reconstruction of 1000+ images in one forward
  pass.
\newblock In \emph{2025 IEEE/CVF Conference on Computer Vision and Pattern
  Recognition (CVPR)}, pp.\  21924--21935. IEEE, 2025.

\bibitem[Ye et~al.(2025)Ye, Liu, Xu, Li, Pollefeys, Yang, and Peng]{nopoSplat}
Botao Ye, Sifei Liu, Haofei Xu, Xueting Li, Marc Pollefeys, Ming-Hsuan Yang,
  and Songyou Peng.
\newblock No pose, no problem: Surprisingly simple 3d gaussian splats from
  sparse unposed images.
\newblock In \emph{International Conference on Learning Representations},
  volume 2025, pp.\  54009--54033, 2025.

\bibitem[Yeshwanth et~al.(2023)Yeshwanth, Liu, Nie{\ss}ner, and Dai]{scannet++}
Chandan Yeshwanth, Yueh-Cheng Liu, Matthias Nie{\ss}ner, and Angela Dai.
\newblock Scannet++: A high-fidelity dataset of 3d indoor scenes.
\newblock In \emph{2023 IEEE/CVF International Conference on Computer Vision
  (ICCV)}, pp.\  12--22. IEEE, 2023.

\bibitem[Zhang et~al.(2024)Zhang, Bi, Tan, Xiangli, Zhao, Sunkavalli, and
  Xu]{gs-lrm}
Kai Zhang, Sai Bi, Hao Tan, Yuanbo Xiangli, Nanxuan Zhao, Kalyan Sunkavalli,
  and Zexiang Xu.
\newblock Gs-lrm: Large reconstruction model for 3d gaussian splatting.
\newblock In \emph{European Conference on Computer Vision}, pp.\  1--19.
  Springer, 2024.

\bibitem[Zhang et~al.(2018)Zhang, Isola, Efros, Shechtman, and Wang]{lpips}
Richard Zhang, Phillip Isola, Alexei~A Efros, Eli Shechtman, and Oliver Wang.
\newblock The unreasonable effectiveness of deep features as a perceptual
  metric.
\newblock In \emph{2018 IEEE/CVF conference on computer vision and pattern
  recognition}, pp.\  586--595. IEEE, 2018.

\bibitem[Zhang et~al.(2025)Zhang, Wang, Xu, Xue, Rupprecht, Zhou, Shen, and
  Wetzstein]{FLARE}
Shangzhan Zhang, Jianyuan Wang, Yinghao Xu, Nan Xue, Christian Rupprecht,
  Xiaowei Zhou, Yujun Shen, and Gordon Wetzstein.
\newblock Flare: Feed-forward geometry, appearance and camera estimation from
  uncalibrated sparse views.
\newblock In \emph{2025 IEEE/CVF Conference on Computer Vision and Pattern
  Recognition (CVPR)}, pp.\  21936--21947. IEEE, 2025.

\bibitem[Zhou et~al.(2018)Zhou, Tucker, Flynn, Fyffe, and Snavely]{re10k}
Tinghui Zhou, Richard Tucker, John Flynn, Graham Fyffe, and Noah Snavely.
\newblock Stereo magnification: Learning view synthesis using multiplane
  images.
\newblock \emph{arXiv preprint arXiv:1805.09817}, 2018.

\bibitem[Ziwen et~al.(2025)Ziwen, Tan, Zhang, Bi, Luan, Hong, Fuxin, and
  Xu]{long-lrm}
Chen Ziwen, Hao Tan, Kai Zhang, Sai Bi, Fujun Luan, Yicong Hong, Li~Fuxin, and
  Zexiang Xu.
\newblock Long-lrm: Long-sequence large reconstruction model for wide-coverage
  gaussian splats.
\newblock In \emph{2025 IEEE/CVF International Conference on Computer Vision
  (ICCV)}, pp.\  4349--4359. IEEE, 2025.

\end{thebibliography}

\clearpage
\appendix
\raggedbottom
\makeatletter
\setlength{\@fptop}{0pt}
\setlength{\@fpsep}{8pt plus 2pt minus 2pt}
\setlength{\@fpbot}{0pt plus 1fil}
\makeatother
\setlength{\parskip}{2pt plus 1pt minus 1pt}
\setlength{\textfloatsep}{10pt plus 2pt minus 2pt}
\setlength{\floatsep}{8pt plus 2pt minus 2pt}
\setlength{\intextsep}{5pt plus 1pt minus 1pt}

\section{Detailed Experimental Setup}
\label{app:implementation}
\suppressfloats[t]

\subsection{Data and View Sampling}
\paragraph{Training data and inputs.}
Both stages sample Hypersim~\citep{hypersim}, Raw3D~\citep{raw3d}, and
ScanNet++~\citep{scannet++} with probabilities of 60\%, 10\%, and 30\%,
using disjoint training and validation scenes. Each sample contains four
target views and $N\in\{4,\ldots,12\}$ input views. The processed image's
longer side is 1022 pixels for $N\leq8$, 756 for $N=9,10$, and 518 for $N=11,12$.

\begin{figure}[t]
\centering
\includegraphics[width=0.78\linewidth]{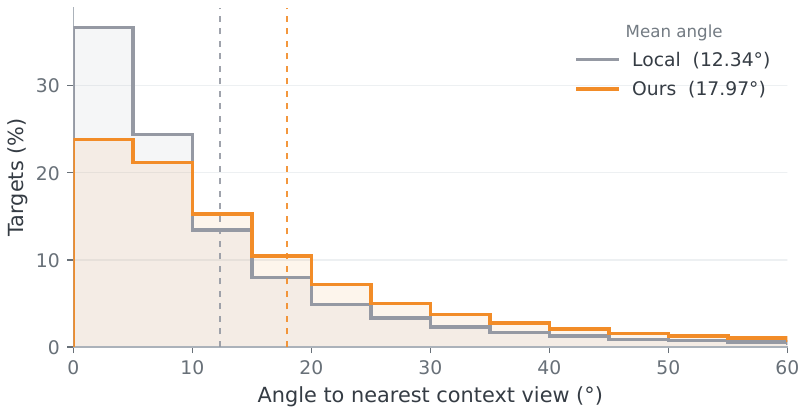}
\caption{Target-view angle distributions. Dashed lines mark the mean minimum
angle to input views: $12.34^\circ$ for local sampling and $17.97^\circ$ for ours.}
\label{fig:target_sampling_angles}
\end{figure}

\paragraph{Target-view sampling.}
\label{app:target_sampling}
We rank non-input views by decreasing minimum optical-axis angle
$a(v,\mathcal C)$ (Eq.~\ref{eq:view_angle_score}), assigning the upper and
lower halves to large- and small-angle pools. Each sample selects the
large-angle pool with probability 0.7 or the small-angle pool with probability
0.3, then draws four target views uniformly without replacement. The selected
pool must contain at least four candidates.
Figure~\ref{fig:target_sampling_angles} shows the sampled angular distributions.

\paragraph{Training masks.}
Target supervision uses visibility masks constructed from GT depth.
A target pixel must have finite GT depth greater than $10^{-6}$ and pass
the supplied validity mask. After back-projection, it is retained if at
least one input view places it in front of the camera and within the
image bounds, has valid sampled depth, and satisfies
\begin{equation}
|z_{\mathrm{proj}}-d_{\mathrm{src}}|
\leq\max(\tau_{\mathrm{abs}},\tau_{\mathrm{rel}}d_{\mathrm{src}}).
\label{eq:warp_depth_consistency}
\end{equation}
Here $z_{\mathrm{proj}}$ is the point depth in that input camera and
$d_{\mathrm{src}}$ is the sampled depth. The mask function defaults to
$\tau_{\mathrm{abs}}=0.03$ and $\tau_{\mathrm{rel}}=0.01$.

\paragraph{Local-view control.}
The w/o V variant samples local interpolation views between input views,
retaining target supervision and the full model's mask construction.

\paragraph{Evaluation views.}
We evaluate on six datasets with $N=4,8,16$ input views.

\subsection{Architecture, Rendering, and Coordinate Conventions}
\paragraph{Initialization and rendering.}
Both branches are initialized from DA3NESTED-GIANT-LARGE-1.1 with independent
parameters. Feature projection and a DPT decoder predict Gaussian attributes.
Voxel merging uses a side length of 0.004 in the scale-normalized reconstruction
frame. During training and inference, Gaussians with opacity below
$\tau_\alpha=0.01$ are pruned after merging and before refinement.
All experiments use gsplat with \texttt{eps2d}=$10^{-2}$.

\paragraph{Scale alignment and point construction.}
\label{app:scale_construction}
Camera alignment applies only a scale factor, not the full similarity transform.
Predicted and GT poses are expressed relative to their respective reference
cameras, using the first input view in sorted order. Predicted depths and
translations are normalized by $s_{\mathrm{norm}}>0$ from
Eq.~\ref{eq:mean_distance_normalization}.
Using only input-camera center correspondences, we fit an Umeyama
similarity transform~\citep{Umeyama} from the GT trajectory to the normalized
predicted trajectory and retain its scale $s_{\mathcal C}$. This scale applies
to both input and target translations, preserving GT intrinsics and
reference-relative rotations:
\begin{equation}
\widetilde{\mathbf R}_v^\star=\mathbf R_v^\star,
\qquad\widetilde{\mathbf t}_v^\star=s_{\mathcal C}\mathbf t_v^\star.
\label{eq:scale_aligned_camera}
\end{equation}
During training, Gaussian centers are constructed from normalized predicted
z-depth, GT intrinsics, and scale-aligned GT poses.

\subsection{Optimization Settings}
Both stages use GT pose conditions and AdamW on 16 H20 GPUs, with
$\beta=(0.9,0.95)$, a global batch size of 16, and global $L_2$ gradient
clipping at 1.0. Each GPU processes one multi-view sample without gradient accumulation.
Stage~I trains only the last ten attribute-backbone Transformer blocks and
the Gaussian head, with peak learning rates of $10^{-6}$ and $2\times10^{-4}$,
respectively. Training runs for 25,000 steps, with 2,000 steps of linear
warmup followed by cosine decay to 0.1 times the peaks, taking about 48 hours.
Stage~II freezes the reconstructor and trains the DPT feature projection
and refiner for 5,000 steps. The learning rate starts at $2\times10^{-4}$
without warmup and decays by a cosine schedule to 0.1 times its initial value,
taking about 13 hours.

\subsection{Stage-I Reconstruction Loss}
\label{app:stage1_loss}
We render the initial scene $\mathcal G^0$ into input and target views
$v\in\mathcal V=\mathcal C\cup\mathcal T$:
\begin{equation}
\widehat{\mathbf I}_v=\mathcal R(\mathcal G^0;\widetilde{\boldsymbol\Pi}_v^\star).
\end{equation}
Let $M_v$ be the valid supervision mask and $e_v(p)$ the RGB-channel-averaged
absolute error at pixel $p$. LPIPS~\citep{lpips} uses VGG features, with invalid
regions zeroed before input normalization, yielding the spatial distance map
\begin{equation}
d_v=\operatorname{LPIPS}_{\mathrm{spatial}}
(M_v\odot\widehat{\mathbf I}_v,\,M_v\odot\mathbf I_v).
\label{eq:image_reconstruction_loss}
\end{equation}
The two losses receive equal weights and are aggregated over valid pixels:
\begin{equation}
\mathcal L_{\mathrm I}
=\frac{\sum_{v\in\mathcal V}\sum_p M_v(p)[e_v(p)+d_v(p)]}
{\max(1,\sum_{v\in\mathcal V}\sum_p M_v(p))}.
\label{eq:stage1_objective}
\end{equation}

\subsection{Stage-II Learned Refinement}
\label{app:refiner_training}
\paragraph{Features and architecture.}
DPT features are projected to eight channels by a learnable $1\times1$
convolution, merged using the Gaussian voxel groups and weights, and reused
across steps. A linear projection initializes the 16-dimensional hidden state.
Image features, current attributes, gradient feedback, neighbor distances,
and hidden states are concatenated and projected to 64 dimensions, then
processed by four KNN Transformer blocks. Each block contains single-head
attention over eight neighbors, LayerNorm, an MLP, and residual connections.
The output is concatenated with the original input, and an MLP predicts
attribute and hidden-state increments.

\paragraph{Feedback and attribute updates.}
During training and inference, all input views provide feedback.
At each step, the current scene is rendered into these views with feedback loss
\begin{equation}
\ell_{\mathrm{fb}}=0.8L_1+0.2L_{\mathrm{DSSIM}},
\qquad L_{\mathrm{DSSIM}}=(1-\mathrm{SSIM})/2.
\end{equation}
DSSIM is computed over valid image regions. Gradients with respect to Gaussian
attributes are normalized and detached before entering the shared updater.
Gaussian centers, counts, and correspondence remain fixed.
Scales update as $s_{\mathrm{new}}=s_{\mathrm{old}}\exp(\Delta\log s)$ without
residual clipping. Opacity-logit, local rotation-vector, and RGB-logit increments
use elementwise $a\tanh(\mathrm{raw})$, with $a=0.1$, $0.25$, and $1.0$,
respectively; hidden-state increments use $0.1\tanh(\mathrm{raw})$.
Logit inputs are clamped to $[10^{-6},1-10^{-6}]$, with no additional cross-step
residual decay. Both stages use zeroth-order SH colors; the refiner updates
colors in RGB-logit coordinates before converting them back to SH coefficients.

\paragraph{Training unroll and supervision.}
Stage~II uses the same input-view counts and resolutions as Stage~I, sampling
$K_{\mathrm r}$ uniformly from $\{1,2,3\}$ updates during training and using
three updates at inference. Input and target views supervise updated scenes.
Feedback gradients and cross-step Gaussian attribute paths are detached,
while hidden-state residual updates remain differentiable for recurrent backpropagation.
The outer loss for each supervised view is
\begin{equation}
\ell_{\mathrm{out}}=0.8L_1+0.2L_{\mathrm{DSSIM}}+0.5L_{\mathrm{LPIPS}}.
\end{equation}
Input and target groups each receive weight 0.5, with equal weights within
each group; invalid views are skipped and remaining weights renormalized.
Let $\mathcal L_{\mathrm{render}}^k$ denote the grouped outer loss after update $k$.
The multi-step objective is
\begin{equation}
\mathcal L_{\mathrm{II}}
=\sum_{k=1}^{K_{\mathrm r}}0.9^{K_{\mathrm r}-k}\mathcal L_{\mathrm{render}}^k
+0.1\mathcal L_{\mathrm{opacity}}.
\label{eq:stage2_objective}
\end{equation}
Step weights are not normalized. Opacity regularization penalizes the change
in logits between final and initial Gaussians:
\begin{equation}
\mathcal L_{\mathrm{opacity}}=\frac1M\sum_{m=1}^{M}
[\operatorname{logit}_{\epsilon}(\alpha_m^{K_{\mathrm r}})
-\operatorname{logit}_{\epsilon}(\alpha_m^0)]^2,
\qquad\epsilon=10^{-6}.
\label{eq:opacity_regularization}
\end{equation}

\paragraph{First-order surrogate loss.}
Let $x^k$ collect updated log-scale, opacity-logit, local-rotation, and RGB-logit
parameters. Rendering gradients are computed on independent leaf copies
$z^k=\operatorname{sg}(x^k)$ to form
\begin{equation}
\begin{aligned}
r^k&=\operatorname{sg}(\nabla_z\mathcal L_{\mathrm{render}}(z^k)),\\
\mathcal L_{\mathrm{proxy}}^k
&=\operatorname{sg}(\mathcal L_{\mathrm{render}}(z^k))
+\langle x^k-\operatorname{sg}(x^k),r^k\rangle.
\end{aligned}
\label{eq:refiner_gradient_proxy}
\end{equation}
Here $\operatorname{sg}$ preserves values while stopping gradients.
The surrogate equals the rendering loss in value, with gradient
\begin{equation}
\nabla_\theta\mathcal L_{\mathrm{proxy}}^k
=\left(\frac{\partial x^k}{\partial\theta}\right)^\top r^k
\label{eq:refiner_proxy_derivative}
\end{equation}
for trainable parameters $\theta$. Per-step surrogates follow the multi-step
objective weights without second-order differentiation.

\paragraph{Chunked execution.}
The global KNN graph is reused across steps. After stable sorting along the
longest voxel-coordinate axis, core sets of at most 200,000 points are expanded
with four-hop neighborhoods. Chunks exceeding 300,000 points are recursively
bisected by point count along their core sets' longest axes, with neighborhoods
recomputed. Only core-point outputs are retained; training uses gradient checkpointing.

\subsection{Camera Protocol and Evaluation Metrics}
\label{app:main_scoring_protocol}
\paragraph{Camera inputs and rendering.}
Pose-conditioned and pose-free results are reported and ranked separately.
In the pose-free setting, reference input cameras are not reconstruction
inputs; they only establish the evaluation reference frame and estimate
translation scale from input-camera correspondences. The scene is constructed
before evaluation-camera alignment and remains unchanged. Target poses define
rendering queries without entering the fit. Alignment rescales reference-relative
translations only, rather than applying a full Sim(3) transform.
For methods predicting intrinsics, target focal lengths are the component-wise
median of predicted input focal lengths, scaled to the target resolution,
with the reference principal point retained. In the pose-conditioned setting,
reference input cameras are available for reconstruction and target intrinsics
are fixed to GT. Evaluation target RGB is never used for reconstruction or refinement feedback.
\paragraph{Metrics and aggregation.}
We evaluate novel view rendering with PSNR, SSIM, LPIPS, and DISTS.
Both settings average equally at each successive level: target views within
each example, examples within each scene, scenes, the three input configurations,
and the six datasets.

\section{Additional Experiments and Results}
\label{app:additional_results}
\subsection{Full Quantitative Results}
\label{app:six_dataset_means}
\paragraph{Six-dataset means.}
Tables~\ref{tab:mean_pose_free} and~\ref{tab:mean_pose_conditioned} summarize
PSNR, SSIM, LPIPS, and DISTS on DL3DV, ETH3D, NRGBD, Mip-NeRF~360, RE10K,
and DTU. Our refined output ranks first on all four mean metrics in the
pose-conditioned setting. In the pose-free setting, refinement gives the best
mean PSNR, SSIM, and LPIPS, while the feed-forward output has the lowest mean DISTS.
\begin{table}[!t]
\centering
\caption{Pose-free results averaged over six datasets.}
\label{tab:mean_pose_free}
\small
\setlength{\tabcolsep}{10pt}
\begin{tabular}{lrrrr}
\toprule
Method & PSNR$\uparrow$ (dB) & SSIM$\uparrow$ & LPIPS$\downarrow$ & DISTS$\downarrow$ \\
\midrule
AnySplat & 15.6668 & 0.5531 & 0.4353 & 0.1939 \\
C3G & 13.2165 & 0.5653 & 0.6454 & 0.3267 \\
QuerySplat & 17.0532 & 0.6404 & 0.4571 & 0.2155 \\
DA3-GS & \cellcolor{nvsthird}17.7144 & \cellcolor{nvssecond}0.6463 & 0.4154 & 0.2077 \\
WorldMirror 2.0 & 17.0031 & 0.6089 & \cellcolor{nvsthird}0.3757 & \cellcolor{nvsthird}0.1935 \\
\midrule
\textbf{Ours} & \cellcolor{nvssecond}18.4002 & \cellcolor{nvsthird}0.6423 & \cellcolor{nvssecond}0.3026 & \cellcolor{nvsfirst}\textbf{0.1341} \\
\textbf{Ours}$^{*}$ & \cellcolor{nvsfirst}\textbf{18.8483} & \cellcolor{nvsfirst}\textbf{0.6523} & \cellcolor{nvsfirst}\textbf{0.2970} & \cellcolor{nvssecond}0.1379 \\
\bottomrule
\end{tabular}
\end{table}

\begin{table}[!t]
\centering
\caption{Pose-conditioned results averaged over six datasets.}
\label{tab:mean_pose_conditioned}
\small
\setlength{\tabcolsep}{10pt}
\begin{tabular}{lrrrr}
\toprule
Method & PSNR$\uparrow$ (dB) & SSIM$\uparrow$ & LPIPS$\downarrow$ & DISTS$\downarrow$ \\
\midrule
tttLRM & 17.8632 & 0.6518 & 0.3793 & 0.2159 \\
DA3-GS & 16.8535 & 0.6100 & 0.4237 & 0.1965 \\
DepthSplat & 18.9085 & 0.6602 & 0.3718 & 0.1993 \\
ReSplat & \cellcolor{nvssecond}20.9139 & \cellcolor{nvssecond}0.7117 & \cellcolor{nvsthird}0.3333 & 0.1794 \\
TokenGS & 15.9149 & 0.5990 & 0.5779 & 0.2931 \\
VolSplat & 14.2476 & 0.5060 & 0.5192 & 0.2756 \\
WorldMirror 2.0 & 17.7222 & 0.6192 & 0.3517 & \cellcolor{nvsthird}0.1734 \\
\midrule
\textbf{Ours} & \cellcolor{nvsthird}20.2824 & \cellcolor{nvsthird}0.6988 & \cellcolor{nvssecond}0.2472 & \cellcolor{nvssecond}0.1163 \\
\textbf{Ours}$^{*}$ & \cellcolor{nvsfirst}\textbf{20.9334} & \cellcolor{nvsfirst}\textbf{0.7131} & \cellcolor{nvsfirst}\textbf{0.2392} & \cellcolor{nvsfirst}\textbf{0.1159} \\
\bottomrule
\end{tabular}
\end{table}

\paragraph{Results by input-view count.}
Tables~\ref{tab:detail_n4}, \ref{tab:detail_n8}, and~\ref{tab:detail_n16}
report per-dataset LPIPS and DISTS with four, eight, and sixteen input views.
All tables follow the colors of Table~\ref{tab:main_nvs}, ranking unrounded
scores within each pose setting and bolding the best; dataset means are equally weighted.
\begin{table}[!p]

\centering
\caption{Per-dataset results with $N=4$ input views. Mean weights datasets equally.}
\label{tab:detail_n4}
\vspace{2pt}
\begingroup
\fontsize{7}{9}\selectfont
\setlength{\tabcolsep}{1.1pt}
\renewcommand{\arraystretch}{1.12}
\resizebox{\linewidth}{!}{%
\begin{tabular}{@{}c@{\hspace{4pt}}l*{14}{c}@{}}
\toprule
\textbf{Input} & \textbf{Method} & \multicolumn{2}{c}{\textbf{DL3DV}} & \multicolumn{2}{c}{\textbf{ETH3D}} & \multicolumn{2}{c}{\textbf{NRGBD}} & \multicolumn{2}{c}{\textbf{Mip-NeRF 360}} & \multicolumn{2}{c}{\textbf{RE10K}} & \multicolumn{2}{c}{\textbf{DTU}} & \multicolumn{2}{c}{\textbf{Mean}} \\
 & & LPIPS$\downarrow$ & DISTS$\downarrow$ & LPIPS$\downarrow$ & DISTS$\downarrow$ & LPIPS$\downarrow$ & DISTS$\downarrow$ & LPIPS$\downarrow$ & DISTS$\downarrow$ & LPIPS$\downarrow$ & DISTS$\downarrow$ & LPIPS$\downarrow$ & DISTS$\downarrow$ & LPIPS$\downarrow$ & DISTS$\downarrow$ \\
\midrule
\multirow{7}{*}{\rotatebox{90}{\textit{Pose-free}}} & AnySplat & 0.6004 & 0.2527 & 0.5244 & 0.2245 & 0.1917 & \cellcolor{nvsthird}0.1153 & 0.5259 & 0.2187 & 0.2386 & \cellcolor{nvsthird}0.1084 & 0.4995 & 0.2452 & 0.4301 & 0.1941 \\
 & C3G & 0.7565 & 0.3785 & 0.6090 & 0.2877 & 0.6137 & 0.3191 & 0.6849 & 0.3448 & 0.5080 & 0.2671 & 0.7031 & 0.3680 & 0.6459 & 0.3275 \\
 & QuerySplat & 0.5453 & 0.2191 & 0.5315 & 0.2425 & 0.2580 & 0.1410 & 0.5621 & 0.2429 & 0.2172 & 0.1117 & 0.5243 & 0.2759 & 0.4397 & 0.2055 \\
 & DA3-GS & 0.5347 & 0.2145 & 0.4937 & 0.2291 & 0.3352 & 0.1703 & 0.4933 & 0.2199 & 0.2412 & 0.1393 & 0.4287 & 0.2411 & 0.4211 & 0.2024 \\
 & WorldMirror 2.0 & \cellcolor{nvsthird}0.4647 & \cellcolor{nvsthird}0.2021 & \cellcolor{nvsthird}0.4484 & \cellcolor{nvsthird}0.2107 & \cellcolor{nvsthird}0.1793 & 0.1386 & \cellcolor{nvsthird}0.4432 & \cellcolor{nvsthird}0.1957 & \cellcolor{nvsthird}0.1903 & 0.1166 & \cellcolor{nvsfirst}\textbf{0.3924} & \cellcolor{nvsthird}0.2076 & \cellcolor{nvsthird}0.3530 & \cellcolor{nvsthird}0.1786 \\
 & \textbf{Ours} & \cellcolor{nvssecond}0.3862 & \cellcolor{nvssecond}0.1322 & \cellcolor{nvssecond}0.3850 & \cellcolor{nvsfirst}\textbf{0.1596} & \cellcolor{nvssecond}0.1078 & \cellcolor{nvssecond}0.0779 & \cellcolor{nvssecond}0.3583 & \cellcolor{nvsfirst}\textbf{0.1390} & \cellcolor{nvssecond}0.1438 & \cellcolor{nvssecond}0.0806 & \cellcolor{nvssecond}0.4162 & \cellcolor{nvsfirst}\textbf{0.1938} & \cellcolor{nvssecond}0.2995 & \cellcolor{nvsfirst}\textbf{0.1305} \\
 & \textbf{Ours}$^{*}$ & \cellcolor{nvsfirst}\textbf{0.3774} & \cellcolor{nvsfirst}\textbf{0.1309} & \cellcolor{nvsfirst}\textbf{0.3718} & \cellcolor{nvssecond}0.1596 & \cellcolor{nvsfirst}\textbf{0.1009} & \cellcolor{nvsfirst}\textbf{0.0765} & \cellcolor{nvsfirst}\textbf{0.3547} & \cellcolor{nvssecond}0.1422 & \cellcolor{nvsfirst}\textbf{0.1416} & \cellcolor{nvsfirst}\textbf{0.0800} & \cellcolor{nvsthird}0.4176 & \cellcolor{nvssecond}0.2029 & \cellcolor{nvsfirst}\textbf{0.2940} & \cellcolor{nvssecond}0.1320 \\
\midrule
\multirow{9}{*}{\rotatebox{90}{\textit{Pose-conditioned}}} & tttLRM & 0.5957 & 0.2962 & 0.5477 & 0.2476 & 0.3496 & 0.2262 & 0.4451 & 0.2349 & 0.2554 & 0.1541 & 0.4803 & 0.3105 & 0.4456 & 0.2449 \\
 & DA3-GS & 0.5265 & 0.2098 & 0.6205 & 0.2595 & 0.3678 & 0.1548 & 0.4851 & 0.2200 & 0.2320 & 0.1248 & 0.3731 & 0.2083 & 0.4342 & 0.1962 \\
 & DepthSplat & 0.5986 & 0.2723 & \cellcolor{nvsthird}0.5085 & 0.2316 & \cellcolor{nvsthird}0.0639 & \cellcolor{nvsthird}0.0550 & 0.3872 & 0.1952 & \cellcolor{nvsfirst}\textbf{0.1066} & \cellcolor{nvsfirst}\textbf{0.0734} & 0.4907 & 0.2622 & 0.3592 & 0.1816 \\
 & ReSplat & \cellcolor{nvsthird}0.4309 & 0.2062 & 0.5511 & 0.2439 & 0.0708 & 0.0755 & \cellcolor{nvsthird}0.3676 & \cellcolor{nvsthird}0.1760 & \cellcolor{nvssecond}0.1073 & 0.0788 & 0.4884 & 0.2923 & \cellcolor{nvsthird}0.3360 & 0.1788 \\
 & TokenGS & 0.7631 & 0.3685 & 0.6764 & 0.2972 & 0.5966 & 0.3103 & 0.6393 & 0.3097 & 0.4131 & 0.2165 & 0.5752 & 0.3505 & 0.6106 & 0.3088 \\
 & VolSplat & 0.6951 & 0.3593 & 0.6539 & 0.3219 & 0.2528 & 0.2053 & 0.4740 & 0.2413 & 0.4248 & 0.2277 & 0.6663 & 0.3312 & 0.5278 & 0.2811 \\
 & WorldMirror 2.0 & 0.4557 & \cellcolor{nvsthird}0.2009 & 0.5264 & \cellcolor{nvsthird}0.2041 & 0.1003 & 0.0705 & 0.4356 & 0.1974 & 0.1886 & 0.1213 & \cellcolor{nvsthird}0.3520 & \cellcolor{nvsthird}0.1954 & 0.3431 & \cellcolor{nvsthird}0.1649 \\
 & \textbf{Ours} & \cellcolor{nvssecond}0.3523 & \cellcolor{nvssecond}0.1259 & \cellcolor{nvssecond}0.4646 & \cellcolor{nvssecond}0.1759 & \cellcolor{nvssecond}0.0634 & \cellcolor{nvssecond}0.0527 & \cellcolor{nvssecond}0.3166 & \cellcolor{nvssecond}0.1387 & 0.1189 & \cellcolor{nvsthird}0.0784 & \cellcolor{nvssecond}0.2226 & \cellcolor{nvsfirst}\textbf{0.1436} & \cellcolor{nvssecond}0.2564 & \cellcolor{nvssecond}0.1192 \\
 & \textbf{Ours}$^{*}$ & \cellcolor{nvsfirst}\textbf{0.3409} & \cellcolor{nvsfirst}\textbf{0.1215} & \cellcolor{nvsfirst}\textbf{0.4462} & \cellcolor{nvsfirst}\textbf{0.1732} & \cellcolor{nvsfirst}\textbf{0.0579} & \cellcolor{nvsfirst}\textbf{0.0499} & \cellcolor{nvsfirst}\textbf{0.3078} & \cellcolor{nvsfirst}\textbf{0.1351} & \cellcolor{nvsthird}0.1134 & \cellcolor{nvssecond}0.0771 & \cellcolor{nvsfirst}\textbf{0.2124} & \cellcolor{nvssecond}0.1448 & \cellcolor{nvsfirst}\textbf{0.2464} & \cellcolor{nvsfirst}\textbf{0.1169} \\
\bottomrule
\end{tabular}
}
\endgroup
\par\vspace{6pt}
\centering
\caption{Per-dataset results with $N=8$ input views. Mean weights datasets equally.}
\label{tab:detail_n8}
\vspace{2pt}
\begingroup
\fontsize{7}{9}\selectfont
\setlength{\tabcolsep}{1.1pt}
\renewcommand{\arraystretch}{1.12}
\resizebox{\linewidth}{!}{%
\begin{tabular}{@{}c@{\hspace{4pt}}l*{14}{c}@{}}
\toprule
\textbf{Input} & \textbf{Method} & \multicolumn{2}{c}{\textbf{DL3DV}} & \multicolumn{2}{c}{\textbf{ETH3D}} & \multicolumn{2}{c}{\textbf{NRGBD}} & \multicolumn{2}{c}{\textbf{Mip-NeRF 360}} & \multicolumn{2}{c}{\textbf{RE10K}} & \multicolumn{2}{c}{\textbf{DTU}} & \multicolumn{2}{c}{\textbf{Mean}} \\
 & & LPIPS$\downarrow$ & DISTS$\downarrow$ & LPIPS$\downarrow$ & DISTS$\downarrow$ & LPIPS$\downarrow$ & DISTS$\downarrow$ & LPIPS$\downarrow$ & DISTS$\downarrow$ & LPIPS$\downarrow$ & DISTS$\downarrow$ & LPIPS$\downarrow$ & DISTS$\downarrow$ & LPIPS$\downarrow$ & DISTS$\downarrow$ \\
\midrule
\multirow{7}{*}{\rotatebox{90}{\textit{Pose-free}}} & AnySplat & 0.5992 & 0.2499 & 0.5586 & 0.2331 & \cellcolor{nvsthird}0.1895 & \cellcolor{nvsthird}0.1128 & 0.5573 & 0.2257 & 0.2486 & 0.1134 & 0.4759 & 0.2332 & 0.4382 & 0.1947 \\
 & C3G & 0.7374 & 0.3521 & 0.7020 & 0.3616 & 0.5210 & 0.2536 & 0.7347 & 0.3706 & 0.4404 & 0.1951 & 0.6952 & 0.3710 & 0.6384 & 0.3173 \\
 & QuerySplat & 0.5581 & 0.2246 & 0.5953 & 0.2840 & 0.2578 & 0.1394 & 0.6096 & 0.2652 & \cellcolor{nvsthird}0.2148 & \cellcolor{nvsthird}0.1106 & 0.5234 & 0.2737 & 0.4599 & 0.2163 \\
 & DA3-GS & 0.5137 & \cellcolor{nvsthird}0.2117 & \cellcolor{nvsthird}0.4897 & 0.2431 & 0.3514 & 0.1794 & 0.4796 & 0.2362 & 0.2592 & 0.1467 & 0.4152 & 0.2337 & 0.4181 & 0.2084 \\
 & WorldMirror 2.0 & \cellcolor{nvsthird}0.4796 & 0.2154 & 0.4908 & \cellcolor{nvsthird}0.2317 & 0.2049 & 0.1536 & \cellcolor{nvsthird}0.4701 & \cellcolor{nvsthird}0.2143 & 0.2218 & 0.1334 & \cellcolor{nvsfirst}\textbf{0.3690} & \cellcolor{nvsthird}0.2074 & \cellcolor{nvsthird}0.3727 & \cellcolor{nvsthird}0.1926 \\
 & \textbf{Ours} & \cellcolor{nvssecond}0.3719 & \cellcolor{nvsfirst}\textbf{0.1276} & \cellcolor{nvssecond}0.4024 & \cellcolor{nvsfirst}\textbf{0.1654} & \cellcolor{nvssecond}0.1101 & \cellcolor{nvssecond}0.0787 & \cellcolor{nvsfirst}\textbf{0.3768} & \cellcolor{nvsfirst}\textbf{0.1482} & \cellcolor{nvssecond}0.1451 & \cellcolor{nvssecond}0.0813 & \cellcolor{nvsthird}0.3782 & \cellcolor{nvsfirst}\textbf{0.1910} & \cellcolor{nvssecond}0.2974 & \cellcolor{nvsfirst}\textbf{0.1320} \\
 & \textbf{Ours}$^{*}$ & \cellcolor{nvsfirst}\textbf{0.3646} & \cellcolor{nvssecond}0.1292 & \cellcolor{nvsfirst}\textbf{0.3899} & \cellcolor{nvssecond}0.1698 & \cellcolor{nvsfirst}\textbf{0.1037} & \cellcolor{nvsfirst}\textbf{0.0777} & \cellcolor{nvssecond}0.3778 & \cellcolor{nvssecond}0.1563 & \cellcolor{nvsfirst}\textbf{0.1439} & \cellcolor{nvsfirst}\textbf{0.0807} & \cellcolor{nvssecond}0.3707 & \cellcolor{nvssecond}0.2036 & \cellcolor{nvsfirst}\textbf{0.2917} & \cellcolor{nvssecond}0.1362 \\
\midrule
\multirow{9}{*}{\rotatebox{90}{\textit{Pose-conditioned}}} & tttLRM & 0.4910 & 0.2639 & \cellcolor{nvsthird}0.5081 & 0.2385 & 0.2522 & 0.1825 & \cellcolor{nvsthird}0.3617 & 0.1922 & 0.2032 & 0.1338 & 0.3605 & 0.2488 & 0.3628 & 0.2099 \\
 & DA3-GS & 0.5156 & \cellcolor{nvsthird}0.2042 & 0.5857 & 0.2505 & 0.3569 & 0.1555 & 0.5018 & 0.2420 & 0.2310 & 0.1264 & 0.3656 & 0.2060 & 0.4261 & 0.1974 \\
 & DepthSplat & 0.5897 & 0.2753 & 0.5814 & 0.3046 & \cellcolor{nvsthird}0.0798 & \cellcolor{nvsthird}0.0656 & 0.4595 & 0.2477 & 0.1261 & \cellcolor{nvsthird}0.0815 & 0.4242 & 0.2381 & 0.3768 & 0.2021 \\
 & ReSplat & \cellcolor{nvsthird}0.4240 & 0.2175 & 0.5472 & 0.2454 & 0.0924 & 0.0887 & 0.3662 & \cellcolor{nvsthird}0.1740 & \cellcolor{nvsthird}0.1214 & 0.0860 & 0.4437 & 0.2732 & \cellcolor{nvsthird}0.3325 & 0.1808 \\
 & TokenGS & 0.6907 & 0.3120 & 0.6730 & 0.2944 & 0.5410 & 0.2987 & 0.6388 & 0.3032 & 0.3896 & 0.2076 & 0.5043 & 0.3189 & 0.5729 & 0.2891 \\
 & VolSplat & 0.7064 & 0.3528 & 0.6813 & 0.3165 & 0.2503 & 0.2035 & 0.4964 & 0.2514 & 0.3848 & 0.2121 & 0.7307 & 0.3831 & 0.5416 & 0.2866 \\
 & WorldMirror 2.0 & 0.4657 & 0.2135 & 0.5273 & \cellcolor{nvsthird}0.2118 & 0.1061 & 0.0710 & 0.4711 & 0.2186 & 0.2175 & 0.1367 & \cellcolor{nvsthird}0.3265 & \cellcolor{nvsthird}0.1921 & 0.3524 & \cellcolor{nvsthird}0.1740 \\
 & \textbf{Ours} & \cellcolor{nvssecond}0.3280 & \cellcolor{nvssecond}0.1191 & \cellcolor{nvssecond}0.4383 & \cellcolor{nvssecond}0.1711 & \cellcolor{nvssecond}0.0656 & \cellcolor{nvssecond}0.0533 & \cellcolor{nvssecond}0.3158 & \cellcolor{nvssecond}0.1374 & \cellcolor{nvssecond}0.1207 & \cellcolor{nvsfirst}\textbf{0.0803} & \cellcolor{nvssecond}0.2041 & \cellcolor{nvsfirst}\textbf{0.1382} & \cellcolor{nvssecond}0.2454 & \cellcolor{nvsfirst}\textbf{0.1166} \\
 & \textbf{Ours}$^{*}$ & \cellcolor{nvsfirst}\textbf{0.3191} & \cellcolor{nvsfirst}\textbf{0.1179} & \cellcolor{nvsfirst}\textbf{0.4216} & \cellcolor{nvsfirst}\textbf{0.1669} & \cellcolor{nvsfirst}\textbf{0.0618} & \cellcolor{nvsfirst}\textbf{0.0529} & \cellcolor{nvsfirst}\textbf{0.3068} & \cellcolor{nvsfirst}\textbf{0.1356} & \cellcolor{nvsfirst}\textbf{0.1184} & \cellcolor{nvssecond}0.0812 & \cellcolor{nvsfirst}\textbf{0.2006} & \cellcolor{nvssecond}0.1455 & \cellcolor{nvsfirst}\textbf{0.2380} & \cellcolor{nvssecond}0.1167 \\
\bottomrule
\end{tabular}
}
\endgroup
\par\vspace{6pt}
\centering
\caption{Per-dataset results with $N=16$ input views. Mean weights datasets equally.}
\label{tab:detail_n16}
\vspace{2pt}
\begingroup
\fontsize{7}{9}\selectfont
\setlength{\tabcolsep}{1.1pt}
\renewcommand{\arraystretch}{1.12}
\resizebox{\linewidth}{!}{%
\begin{tabular}{@{}c@{\hspace{4pt}}l*{14}{c}@{}}
\toprule
\textbf{Input} & \textbf{Method} & \multicolumn{2}{c}{\textbf{DL3DV}} & \multicolumn{2}{c}{\textbf{ETH3D}} & \multicolumn{2}{c}{\textbf{NRGBD}} & \multicolumn{2}{c}{\textbf{Mip-NeRF 360}} & \multicolumn{2}{c}{\textbf{RE10K}} & \multicolumn{2}{c}{\textbf{DTU}} & \multicolumn{2}{c}{\textbf{Mean}} \\
 & & LPIPS$\downarrow$ & DISTS$\downarrow$ & LPIPS$\downarrow$ & DISTS$\downarrow$ & LPIPS$\downarrow$ & DISTS$\downarrow$ & LPIPS$\downarrow$ & DISTS$\downarrow$ & LPIPS$\downarrow$ & DISTS$\downarrow$ & LPIPS$\downarrow$ & DISTS$\downarrow$ & LPIPS$\downarrow$ & DISTS$\downarrow$ \\
\midrule
\multirow{7}{*}{\rotatebox{90}{\textit{Pose-free}}} & AnySplat & 0.5958 & 0.2445 & 0.5588 & \cellcolor{nvsthird}0.2315 & \cellcolor{nvsthird}0.1873 & \cellcolor{nvsthird}0.1112 & 0.5705 & \cellcolor{nvsthird}0.2334 & 0.2490 & 0.1120 & 0.4646 & 0.2251 & 0.4377 & \cellcolor{nvsthird}0.1929 \\
 & C3G & 0.7571 & 0.3775 & 0.7056 & 0.3921 & 0.4833 & 0.2398 & 0.7973 & 0.4479 & 0.4315 & 0.1813 & 0.7362 & 0.3722 & 0.6518 & 0.3351 \\
 & QuerySplat & 0.5769 & 0.2342 & 0.6146 & 0.2986 & 0.2599 & 0.1411 & 0.6510 & 0.2954 & \cellcolor{nvsthird}0.2156 & \cellcolor{nvsthird}0.1116 & 0.5121 & 0.2669 & 0.4717 & 0.2246 \\
 & DA3-GS & \cellcolor{nvsthird}0.4953 & 0.2276 & \cellcolor{nvsthird}0.5017 & 0.2581 & 0.3501 & 0.1835 & \cellcolor{nvsthird}0.4483 & 0.2350 & 0.2574 & 0.1441 & 0.3885 & 0.2262 & 0.4069 & 0.2124 \\
 & WorldMirror 2.0 & 0.4956 & \cellcolor{nvsthird}0.2269 & 0.5360 & 0.2506 & 0.2457 & 0.1746 & 0.4971 & 0.2408 & 0.2511 & 0.1470 & \cellcolor{nvsthird}0.3821 & \cellcolor{nvsthird}0.2168 & \cellcolor{nvsthird}0.4013 & 0.2095 \\
 & \textbf{Ours} & \cellcolor{nvssecond}0.3736 & \cellcolor{nvsfirst}\textbf{0.1280} & \cellcolor{nvssecond}0.4560 & \cellcolor{nvsfirst}\textbf{0.1881} & \cellcolor{nvssecond}0.1109 & \cellcolor{nvssecond}0.0799 & \cellcolor{nvsfirst}\textbf{0.4000} & \cellcolor{nvsfirst}\textbf{0.1641} & \cellcolor{nvssecond}0.1489 & \cellcolor{nvssecond}0.0838 & \cellcolor{nvssecond}0.3751 & \cellcolor{nvsfirst}\textbf{0.1948} & \cellcolor{nvssecond}0.3107 & \cellcolor{nvsfirst}\textbf{0.1398} \\
 & \textbf{Ours}$^{*}$ & \cellcolor{nvsfirst}\textbf{0.3673} & \cellcolor{nvssecond}0.1318 & \cellcolor{nvsfirst}\textbf{0.4444} & \cellcolor{nvssecond}0.1919 & \cellcolor{nvsfirst}\textbf{0.1055} & \cellcolor{nvsfirst}\textbf{0.0787} & \cellcolor{nvssecond}0.4038 & \cellcolor{nvssecond}0.1799 & \cellcolor{nvsfirst}\textbf{0.1480} & \cellcolor{nvsfirst}\textbf{0.0818} & \cellcolor{nvsfirst}\textbf{0.3623} & \cellcolor{nvssecond}0.2079 & \cellcolor{nvsfirst}\textbf{0.3052} & \cellcolor{nvssecond}0.1454 \\
\midrule
\multirow{9}{*}{\rotatebox{90}{\textit{Pose-conditioned}}} & tttLRM & \cellcolor{nvsthird}0.3998 & \cellcolor{nvsthird}0.2123 & \cellcolor{nvsthird}0.4395 & \cellcolor{nvsthird}0.2027 & 0.2554 & 0.1850 & \cellcolor{nvsthird}0.3437 & 0.1837 & 0.1993 & 0.1333 & 0.3392 & 0.2393 & \cellcolor{nvsthird}0.3295 & 0.1927 \\
 & DA3-GS & 0.5284 & 0.2274 & 0.5267 & 0.2228 & 0.3531 & 0.1569 & 0.5025 & 0.2550 & 0.2316 & 0.1276 & 0.3220 & \cellcolor{nvsthird}0.1858 & 0.4107 & 0.1959 \\
 & DepthSplat & 0.5413 & 0.2646 & 0.5945 & 0.3260 & 0.1163 & 0.0814 & 0.5044 & 0.2915 & 0.1563 & 0.0972 & 0.3640 & 0.2235 & 0.3795 & 0.2140 \\
 & ReSplat & 0.4043 & 0.2133 & 0.4815 & 0.2034 & 0.1980 & 0.1245 & 0.3627 & \cellcolor{nvsthird}0.1741 & \cellcolor{nvsthird}0.1287 & \cellcolor{nvsthird}0.0877 & 0.4133 & 0.2687 & 0.3314 & \cellcolor{nvsthird}0.1786 \\
 & TokenGS & 0.6196 & 0.2913 & 0.6618 & 0.2822 & 0.5392 & 0.2984 & 0.6586 & 0.3155 & 0.3838 & 0.2043 & 0.4377 & 0.2969 & 0.5501 & 0.2814 \\
 & VolSplat & 0.7132 & 0.3495 & 0.6792 & 0.3064 & 0.1996 & 0.1644 & 0.5509 & 0.2826 & 0.2560 & 0.1558 & 0.5301 & 0.2966 & 0.4882 & 0.2592 \\
 & WorldMirror 2.0 & 0.4805 & 0.2234 & 0.5066 & 0.2078 & \cellcolor{nvsthird}0.1104 & \cellcolor{nvsthird}0.0735 & 0.5001 & 0.2423 & 0.2461 & 0.1496 & \cellcolor{nvsthird}0.3134 & 0.1914 & 0.3595 & 0.1813 \\
 & \textbf{Ours} & \cellcolor{nvssecond}0.3285 & \cellcolor{nvsfirst}\textbf{0.1176} & \cellcolor{nvssecond}0.3989 & \cellcolor{nvssecond}0.1514 & \cellcolor{nvssecond}0.0669 & \cellcolor{nvssecond}0.0544 & \cellcolor{nvssecond}0.3292 & \cellcolor{nvssecond}0.1453 & \cellcolor{nvssecond}0.1211 & \cellcolor{nvsfirst}\textbf{0.0808} & \cellcolor{nvssecond}0.1942 & \cellcolor{nvsfirst}\textbf{0.1287} & \cellcolor{nvssecond}0.2398 & \cellcolor{nvsfirst}\textbf{0.1130} \\
 & \textbf{Ours}$^{*}$ & \cellcolor{nvsfirst}\textbf{0.3227} & \cellcolor{nvssecond}0.1196 & \cellcolor{nvsfirst}\textbf{0.3850} & \cellcolor{nvsfirst}\textbf{0.1491} & \cellcolor{nvsfirst}\textbf{0.0640} & \cellcolor{nvsfirst}\textbf{0.0543} & \cellcolor{nvsfirst}\textbf{0.3150} & \cellcolor{nvsfirst}\textbf{0.1433} & \cellcolor{nvsfirst}\textbf{0.1210} & \cellcolor{nvssecond}0.0823 & \cellcolor{nvsfirst}\textbf{0.1911} & \cellcolor{nvssecond}0.1362 & \cellcolor{nvsfirst}\textbf{0.2331} & \cellcolor{nvssecond}0.1141 \\
\bottomrule
\end{tabular}
}
\endgroup

\end{table}

\subsection{Feed-Forward Reconstructor Ablations}
\label{app:reconstructor_ablation}
\begin{figure}[!t]
\centering
\begin{minipage}{0.6\linewidth}
\input{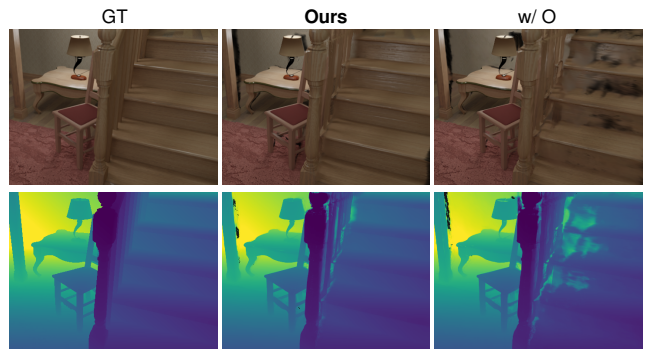}
\end{minipage}
\caption{Positional offsets on NRGBD with eight input views.
Top: RGB; bottom: depth. Rendered depth is alpha-normalized expected camera-$z$
depth, without fitting to target GT. Colors indicate near (purple), far (yellow),
and invalid (gray) pixels.}
\label{fig:offset_additional}
\end{figure}

All variants are evaluated on ETH3D and DTU before refinement, averaging
equally over $N=4,8,16$, with other training settings unchanged.
The w/o A variant freezes the attribute backbone ($K=0$), shares backbone
features across branches, and trains only the Gaussian head. The w/o G variant
fine-tunes the geometry backbone at $10^{-6}$, keeping the Gaussian head learning
rate at $2\times10^{-4}$. The w/ O variant keeps the geometry backbone frozen
and predicts 3D offsets added to back-projected centers in the scale-aligned
frame. The w/o C variant uses predicted cameras for training-time back-projection.
The w/o V variant uses local-view sampling, retaining target supervision and
the same training masks.
Figure~\ref{fig:offset_additional} provides an NRGBD staircase comparison,
where the offset variant shows stronger surface artifacts and depth deviations
than the fixed-center model.

\subsection{Learned Refinement versus Direct Gradient Descent}
\label{app:refiner_ablation}
We compare feed-forward outputs, Adam, and learned refinement on ETH3D,
averaging equally over $N=4,8,16$. Both refinements start from the same scene
and perform three updates to scales, opacities, rotations, and colors with
centers fixed. Adam uses a learning rate of 0.01.
\begin{table}[!t]
\centering
\caption{Three-step refinement on ETH3D, averaged over $N=4,8,16$ input views.}
\label{tab:ablation_refiner_v2}
\small
\setlength{\tabcolsep}{5pt}
\begin{tabular}{lcccc}
\toprule
Variant & PSNR$\uparrow$ (dB) & SSIM$\uparrow$ & LPIPS$\downarrow$ & DISTS$\downarrow$ \\
\midrule
Ours & \cellcolor{nvsthird}16.9643 & \cellcolor{nvsthird}0.5836 & \cellcolor{nvsthird}0.4339 & \cellcolor{nvsthird}0.1661 \\
Ours + Adam & \cellcolor{nvssecond}16.9890 & \cellcolor{nvssecond}0.5864 & \cellcolor{nvssecond}0.4312 & \cellcolor{nvssecond}0.1656 \\
Ours$^{*}$ & \cellcolor{nvsfirst}\textbf{17.6074} & \cellcolor{nvsfirst}\textbf{0.6067} & \cellcolor{nvsfirst}\textbf{0.4176} & \cellcolor{nvsfirst}\textbf{0.1631} \\
\bottomrule
\end{tabular}
\end{table}

\begin{figure}[!t]
\centering
\input{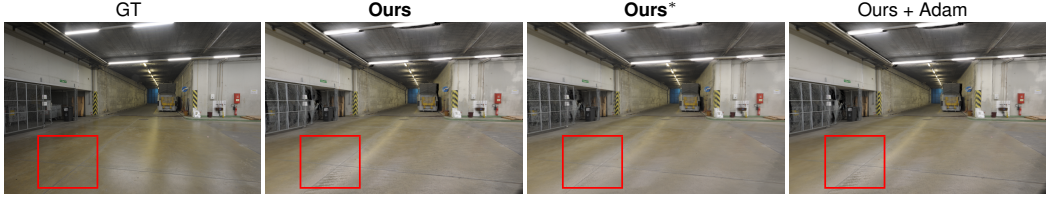}
\caption{Three-step refinement on ETH3D with eight input views.
Red boxes highlight the compared floor region.}
\label{fig:refiner_delivery}
\end{figure}

Learned refinement and Adam improve PSNR by 0.64 dB and 0.02 dB, respectively;
the learned refiner also performs better on the other three metrics.
Figure~\ref{fig:refiner_delivery} provides a qualitative comparison.

\subsection{Opacity and Geometric Consistency}
\label{app:opacity_diagnostic}
We analyze feed-forward outputs on held-out Hypersim~\citep{hypersim}
validation scenes. Geometric labels are assigned independently of opacity
using GT depth, with tolerance $\max(0.03\,\mathrm m,0.01d)$ and depth-edge
samples excluded.
A point is geometrically inconsistent if it violates GT-observed free space
in at least two views and agrees with no observed surface depth.
Surface-consistent points agree with at least two views and have no free-space
violations; others remain uncertain. This identifies 36,125 inconsistent points.
Opacity distributions and pruning statistics at threshold 0.01 are shown in
Fig.~\ref{fig:opacity_diagnostic}.

\subsection{Input Resolution and Feed-Forward Network Latency}
\label{app:resolution_latency}
We measure feed-forward neural-module latency at four resolutions on one H20,
using the same ETH3D scene and eight input views. Each resolution uses one
warmup followed by five synchronized CUDA-event measurements.
\begin{table}[!t]
\centering
\caption{Feed-forward network latency: mean and min--max over five runs.}
\label{tab:resolution-time}
\small
\begin{tabular}{lrr}
\toprule
Input resolution ($W\times H$) & Mean (s) & Min--max (s) \\
\midrule
1022$\times$756 & 5.619 & 5.617--5.621 \\
756$\times$560 & 2.533 & 2.532--2.535 \\
504$\times$378 & 1.032 & 1.031--1.034 \\
378$\times$280 & 0.564 & 0.563--0.564 \\
\bottomrule
\end{tabular}
\end{table}

\subsection{Additional Qualitative Results}
\label{app:additional_qualitative}
\begin{figure}[!t]
\centering
\begingroup
\centering
\setlength{\tabcolsep}{1pt}
\begin{tabular}{@{}cccccc@{}}
{\scriptsize\sffamily WorldMirror 2.0} & {\scriptsize\sffamily AnySplat} & {\scriptsize\sffamily QuerySplat} & {\scriptsize\sffamily DA3-GS} & {\scriptsize\sffamily Ours} & {\scriptsize\sffamily GT} \\[2pt]
\includegraphics[width=0.16\linewidth]{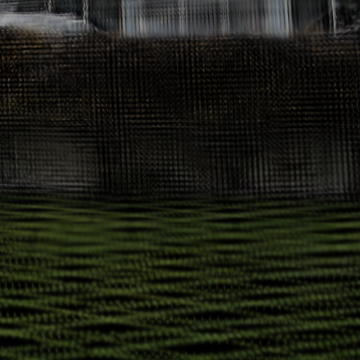} & \includegraphics[width=0.16\linewidth]{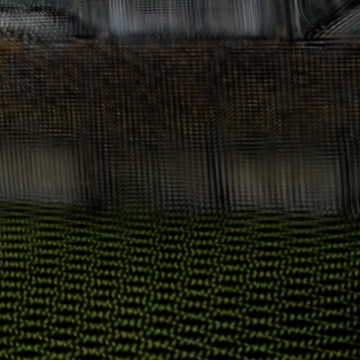} & \includegraphics[width=0.16\linewidth]{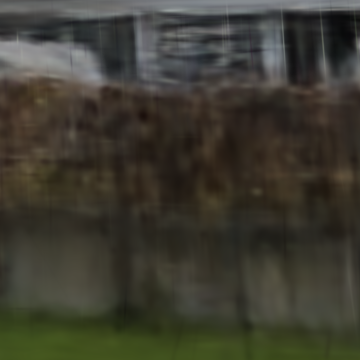} & \includegraphics[width=0.16\linewidth]{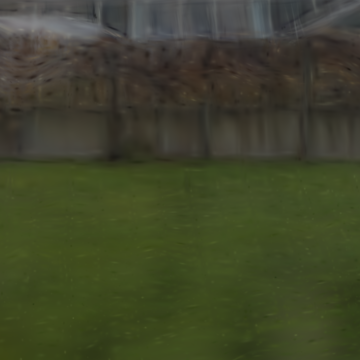} & \includegraphics[width=0.16\linewidth]{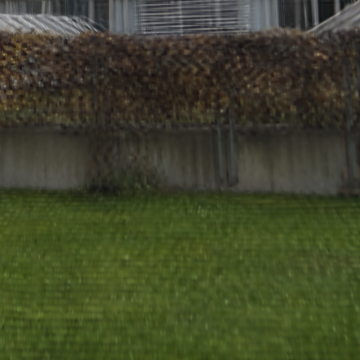} & \includegraphics[width=0.16\linewidth]{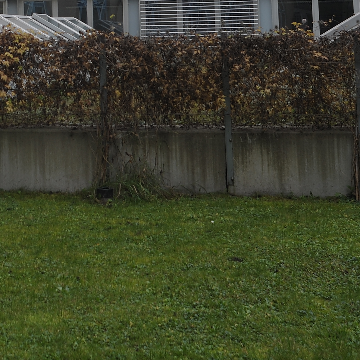} \\[2pt]
\includegraphics[width=0.16\linewidth]{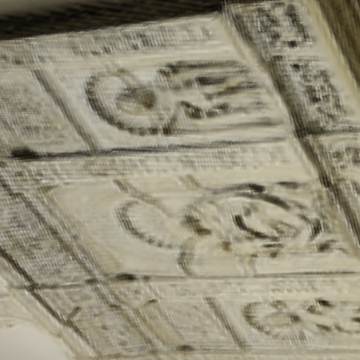} & \includegraphics[width=0.16\linewidth]{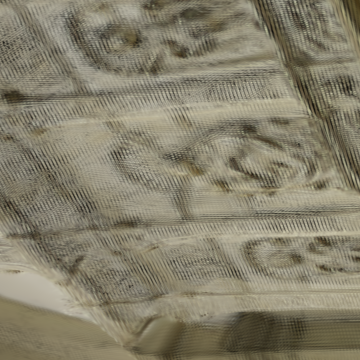} & \includegraphics[width=0.16\linewidth]{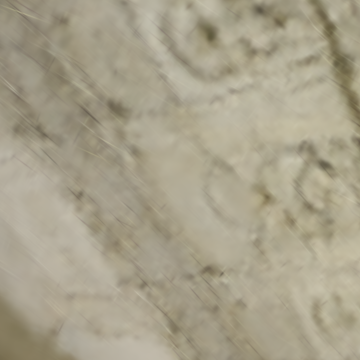} & \includegraphics[width=0.16\linewidth]{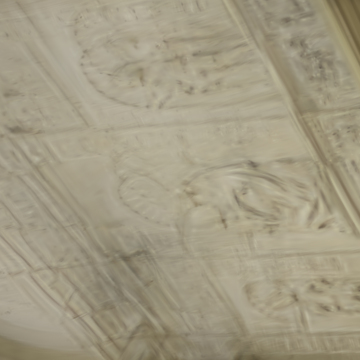} & \includegraphics[width=0.16\linewidth]{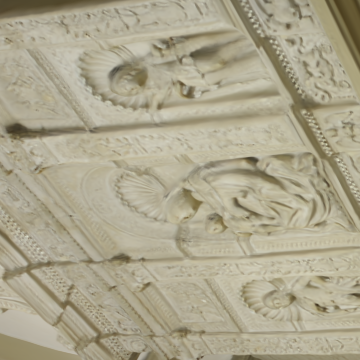} & \includegraphics[width=0.16\linewidth]{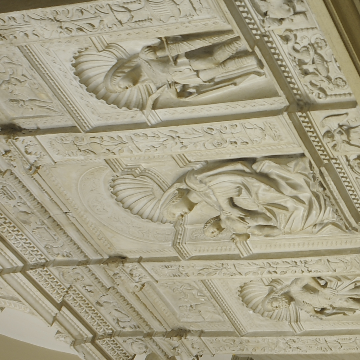} \\[2pt]
\end{tabular}
\endgroup

\caption{Pose-free rendering on ETH3D: grass textures and relief details.}
\label{fig:pose_free_eth3d}
\end{figure}

\paragraph{Pose-free comparisons.}
\label{app:pose_free_qualitative}
Figure~\ref{fig:pose_free_eth3d} compares pose-free reconstructions on ETH3D.
The selected crops show fewer repetitive stripe artifacts and better-preserved
relief structures than several baselines.

\begin{figure}[!t]
\centering
\includegraphics[width=\linewidth]{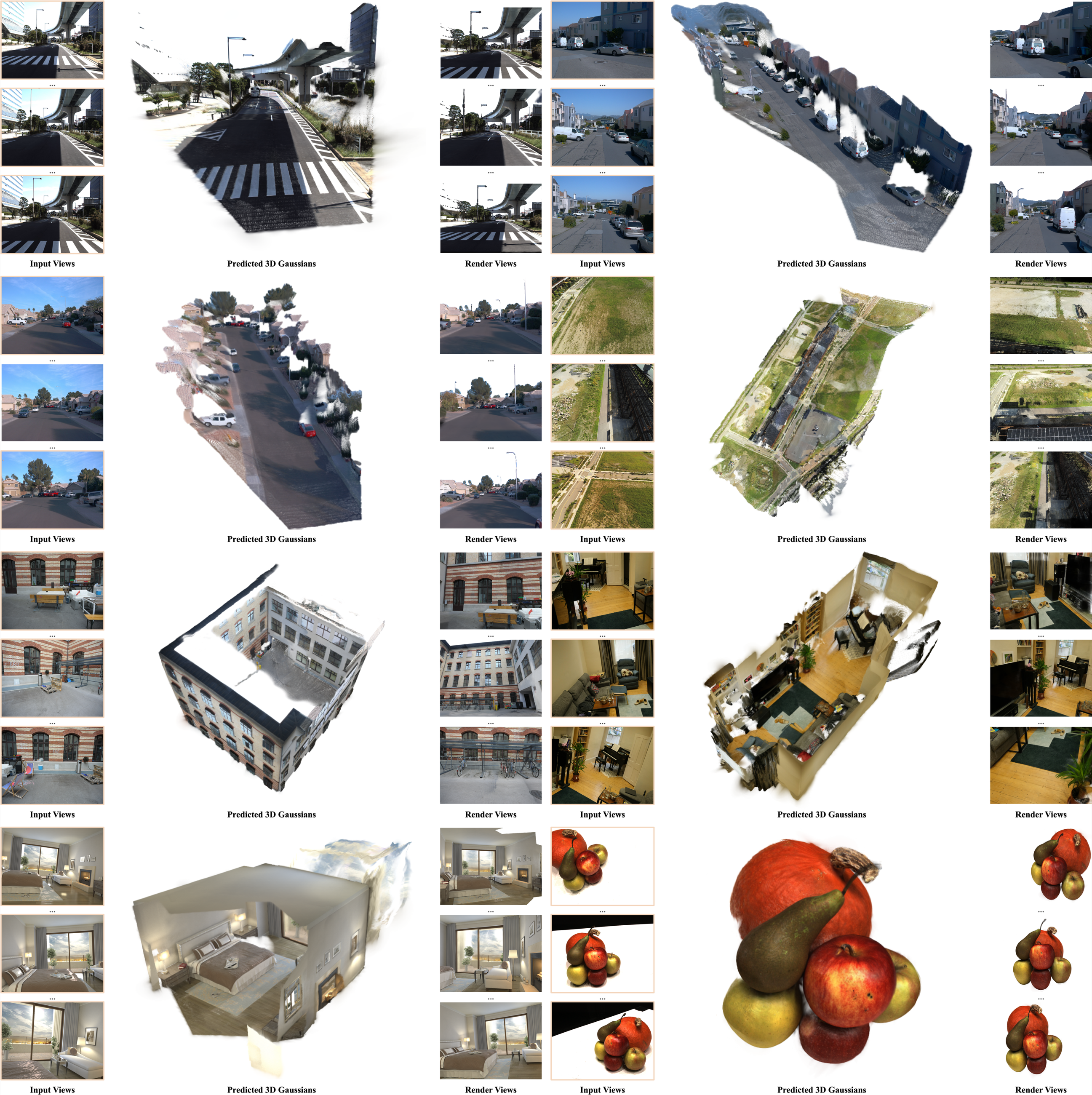}
\caption{Feed-forward reconstructions and renderings with 12--64 input views.}
\label{fig:additional_qualitative}
\end{figure}

\paragraph{Diverse scenes.}
Figure~\ref{fig:additional_qualitative} shows feed-forward reconstructions on
DDAD~\citep{ddad}, Waymo~\citep{waymo}, ETH3D~\citep{eth3d},
Hypersim~\citep{hypersim}, DTU~\citep{DTU}, Mip-NeRF~360~\citep{mip-nerf},
and Mill19 (Mega-NeRF)~\citep{mega-nerf}, covering streets, buildings,
interiors, and objects.

\begin{figure}[!t]
\centering
\includegraphics[width=\linewidth]{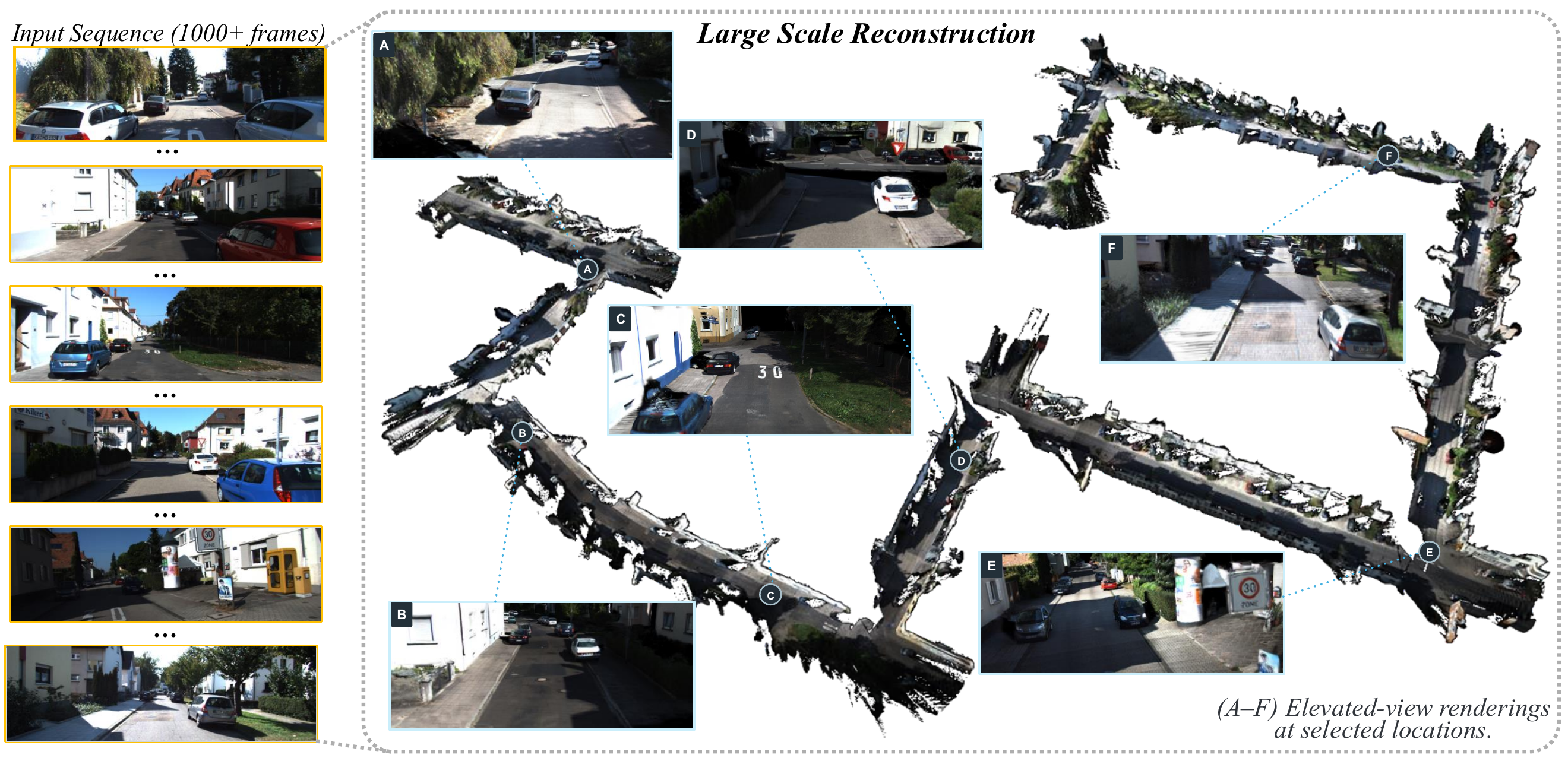}
\caption{Large-scale reconstruction on KITTI.
Insets show elevated-view renderings at locations A--F.}
\label{fig:application_kitti}
\end{figure}

\paragraph{Large-scale reconstruction.}
\label{sec:exp_application}
We apply the feed-forward reconstructor to a KITTI~\citep{KITTI} street
sequence of over 1,000 frames using sliding windows to construct a large-scale
Gaussian scene (Fig.~\ref{fig:application_kitti}).

\Needspace{11\baselineskip}
\section{Limitations and Future Work}
\label{app:limitations}
Opacity learning and pruning can suppress unreliable Gaussians, but fixed
centers cannot correct severe errors in predicted geometry. Reconstruction
quality therefore remains limited by the geometric prior. Pixel-aligned
prediction also produces redundancy not fully removed by voxel merging,
while limited input coverage can leave holes in unobserved regions.
Future work will explore stronger geometric priors and more compact Gaussian
representations, as well as generative models and camera-trajectory design
to address limited coverage.

\end{document}